\documentclass[10pt,journal,compsoc]{IEEEtran}
\ifCLASSOPTIONcompsoc
  \usepackage[nocompress]{cite}
\else
  \usepackage{cite}
\fi

\ifCLASSINFOpdf
\else
\fi
\usepackage{array}

\usepackage{graphicx}
\usepackage{amssymb}
\usepackage{comment}
\usepackage{amsmath}
\usepackage{epstopdf}
\usepackage{algorithm}
\usepackage{algorithmicx}
\usepackage{algpseudocode}
\usepackage{mathrsfs} 
\DeclareMathOperator*{\argmax}{argmax}
\DeclareMathOperator*{\argmin}{argmin}
\usepackage{multirow}
\usepackage{subfigure}
\usepackage{color}
\usepackage{bbm}
\usepackage{url}
\usepackage{booktabs}
\begin{document}
%
\title{GrCAN: Gradient Boost Convolutional Autoencoder with Neural Decision Forest}
%
%
%
%

\author{Manqing~Dong, 
        Lina~Yao, 
        Xianzhi~Wang, 
        Boualem~Benatallah, 
        and Shuai Zhang
\IEEEcompsocitemizethanks{\IEEEcompsocthanksitem M. Dong, L. Yao, X. Wang, B. Bentallah and S. Zhang are with the Department
of Computer Science and Engineering, University of New South Wales, Sydney,
Australia.}
\thanks{Manuscript received April 19, 2005; revised August 26, 2015.}}

%
%

\markboth{Journal of \LaTeX\ Class Files,~Vol.~14, No.~8, August~2015}%
{Shell \MakeLowercase{\textit{et al.}}: GrCAN: Gradient Boost Convolutional Autoencoder with Neural Decision Forest}
%



\IEEEtitleabstractindextext{%
\begin{abstract}

Random forest and deep neural network are two schools of effective classification methods in machine learning. While the random forest is robust irrespective of the data domain, deep neural network has advantages in handling high dimensional data. In view that a differentiable neural decision forest can be added to the neural network to fully exploit the benefits of both models, in our work, we further combine convolutional autoencoder with neural decision forest, where autoencoder has its advantages in finding the hidden representations of the input data. We develop a gradient boost module and embed it into the proposed convolutional autoencoder with neural decision forest to improve the performance. The idea of gradient boost is to learn and use the residual in the prediction. In addition, we design a structure to learn the parameters of the neural decision forest and gradient boost module at contiguous steps. The extensive experiments on several public datasets demonstrate that our proposed model achieves good efficiency and prediction performance compared with a series of baseline methods. 

\end{abstract}

}

\maketitle

\IEEEdisplaynontitleabstractindextext

%
\IEEEpeerreviewmaketitle

\IEEEraisesectionheading{\section{Introduction}\label{sec:introduction}}

%
%
%
%

 

\IEEEPARstart{M}{achine} learning techniques have shown great power and efficacy in dealing with various tasks in the past few decades\cite{liu2017survey}. Among existing machine learning models, Random forest\cite{breiman2001random} and deep neural network\cite{lecun2015deep} are two promising classes of methods that have been proven successful in many applications. 
%
Random forest is the ensemble of decision trees\cite{breiman2001random}. It can alleviate the possible overfitting of decision trees to their training set\cite{biau2016random} and therefore is robust irrespective of the data domain \cite{fernandez2014do}. In past decades, random forest has been successfully applied in various practical problems, such as 3D object recognition\cite{shotton2013real} and economics\cite{varian2014big}. 
On the other hand, deep neural networks recently have been revolutionizing many domain applications, such as speech recognition\cite{graves2013speech}, computer vision tasks\cite{he2016deep}, and a more specific example: Alphago \cite{silver2016mastering}, which first won human players in chess. As more and more works are conducted with the neural network (NN), neural network is becoming bigger, deeper, and more complicated. Taking the champions of ImageNet Large Scale Visual Recognition Challenge (ILSVRC) for example, there are nine layers in the neural network model, AlexNet, in 2012\cite{krizhevsky2012imagenet}, and this number is increased to 152 in the best model, ResNet, in 2015's contest\cite{he2016deep}. 

Despite the advantages, neither of these two models can serve as the one solution for all: while neural networks show power in dealing with large-scale data, this power degrades with the increasing risk of over-fitting over smaller datasets\cite{zhou2017deep}; likewise, while random forest is suitable for many occasions, it often yields sub-optimal classification performance due to the greedy tree construction process \cite{biau2016random}. Several researchers try to combine the strength of both models \cite{manqing2018, welbl2014casting}. For example, in view that traditional random forest is not differentiable, researchers try to transform it to a differentiable one and add it into neural networks. A typical work is porposed by Johannes et. al \cite{welbl2014casting}, who point out that any decision trees can be represented as a two-layer Convolutional Neural Network (CNN)\cite{krizhevsky2012imagenet}, where the first layer includes the decision nodes and the second layer leaf nodes. They then implement a cast random forest, which performs better than traditional random forest and neural network over several UCI datasets. Apart from Johannes' work, Peter Kontschieder et. al \cite{kontschieder2015deep} introduce statistics into the neural network. In particular, they represent decision nodes as probability distributions to make the whole decision differentiable. By combining the designed neural decision forest with CNN, they finally achieve good results on two benchmark image recognition datasets. 

Apart from combining random forest with neural networks, the local minima problem, which many classification problems are deducted to, is able to be dealt with using ensembles of models \cite{hansen1990neural,friedman2001greedy}. Since random forest could be regarded as the bagging version of decision trees, a boosting version of the decision tree, i.e., gradient boost decision tree (GBDT)\cite{friedman2001greedy}, improves the decision tree and is another powerful machine learning tool beside the random forest. For neural networks, it is proved that ensembles of networks can alleviate the local minima situation \cite{hansen1990neural}. First attempts include Schwenk \cite{schwenk2000boosting}'s work, which fuses the Adaboost\cite{collins2002logistic} idea into neural networks.

Inspired by the previous work, we combine them all to propose GrCAN, short for Gradient boost Convolutional Autoencoder with Neural decision forest. To the best of our knowledge, no previous work has introduced the gradient boost idea into the neural decision forest. In particular, we bring gradient boost modules into the neural decision forest and present an end-to-end unified model to leverage the appealing properties of both neural decision forest and gradient boost. We also leverage traditional deep learning module, convolutional autoencoder\cite{chen2017deep}, with the neural decision forest. Since autoencoder\cite{lecun2015deep} is an unsupervised model for learning the hidden representations of the input data, adding CNN enables it to capture the relationships of the neighborhoods in inputs. We further present the strategy of integrating multiple gradient boost modules into the neural decision forest and conduct extensive experiments over three public datasets. Our results show the proposed model is efficient and general enough to solve a wide range of problems, from classification problem for a small dataset to high-dimensional image recognition and text analysis tasks. In a nutshell, we make the following contributions:

\begin{itemize}
\item We bring the gradient boost idea into neural decision forest and unify them in an end-to-end trainable way. 
\item We further make the gradient boost modules extendable to enhance the model's robustness and performance.
\item We evaluate the proposed model on different kind and size of the datasets and achieve superior prediction performance compared with a series of baseline and state-of-the-art methods. 
\end{itemize}

\section{Related Work}
Both Random Forest classifiers\cite{breiman2001random} and feed forward Artificial Neural Networks (ANNs)\cite{schalkoff1997artificial} are widely used supervised machine learning methods for classification. 

Random forest algorithm is an ensemble learning method which normally deals with classification or regression tasks\cite{breiman2001random}. The model has been modified for several versions after it was proposed. The random forest that we usually talk about is the one that combines several randomized decision trees and aggregates their predictions by averaging\cite{ho1995random}. The two most important ingredients of the random forest are the bagging\cite{breiman2001random} and CART(Classification And Regression Trees (CART)-split criterion) split criterion\cite{breiman1984classification}. Bagging is a general aggregation method, it could construct a predictor from each sample, and take the average prediction for the decision. It's said\cite{biau2016random} it is one of the most effective computationally intensive procedures to improve on unstable estimates, especially for large, high-dimensional data sets. Random forests has many interesting properties, such as it will show good performance when a larger number of variables than the number of observations is set\cite{biau2016random}. Lin et. al. found there is some relationship between random forest and nearest neighbor algorithms. And scornet et al. \cite{scornet2015consistency} found that random forest is consistent in an additive regression framework. 

The concept of deep learning originated from the study on artificial neural networks (ANNs) \cite{schalkoff1997artificial}. With the rapid development of computation techniques, a powerful framework has been provided by ANNs with deep architectures for supervised learning, where deep learning was introduced\cite{liu2017survey}. And milestone of deep learning could be traced from 2006 when Hinton proposed a novel deep structured learning architecture called deep belief network (DBN) \cite{hinton2006fast}. Normally we call a neural network which contains three or more layers deep neural networks (DNNs). And in recent years, neural networks layers has become deeper and with higher complexity. Besides, a variety of structures for DNNs were proposed in the past 10 years. One of the most famous ones is called restricted Boltzmann machines (RBMs)\cite{hinton2012practical}, which could learn the probability distribution with the inputs for each module. We could regard an RBM as a special type of Markov random fields\cite{liu2017survey}. Convolutional Neural Networks (CNN) and Recurrent Neural Networks (RNNs)\cite{lecun2015deep} are two another widely used deep learning methods. Especially in recent years, they got lots of achievements in dealing with image recognition\cite{simonyan2014very}, activity recognition\cite{ordonez2016deep}, document classification\cite{song2017hierarchical}, and many other new tasks \cite{dalinAAAI18}. While CNN is good at dealing images information, RNNs is good at discovering query traces. Another famous deep learning model is Autoencoder (AE)\cite{lecun2015deep}, which learns the hidden representation of the input data, and typically for the purpose of dimension reduction.

Despite their advantages, both random forest and deep neural networks have their limitations. The random forest is prone to overfitting while handling noisy data \cite{biau2016random}, and deep neural networks usually have many hyper-parameters, making the performance depends heavily on the parameter tuning \cite{zhou2017deep}.
For the above reasons, some research seeks to bridge the two model for leveraging the benefits of both of them. One of the most important work in this area is the studies that cast the traditional random forest as the structure of neural networks. The first few attempts started in the 1990s when Sethi\cite{sethi1990entropy} proposed entropy net, which encoded decision trees into neural networks. Welbl et. al \cite{welbl2014casting} further proposed that any decision trees can be represented as two-layer Convolutional Neural Networks (CNN).
Since the classical random forest cannot conduct back-propagation, other researchers tried to design differentiable decision tree methods, which can be traced from Montillo's work \cite{montillo2013entanglement}. In this work, they investigated the use of sigmoidal functions for the task of differentiable information gain maximization. In Peter et al.'s work\cite{kontschieder2015deep}, they introduce a stochastic, differentiable, back propagation compatible version of decision trees, guiding the representation learning in lower layers of deep convolutional networks. They implemented their model over several benchmark image recognition datasets and beats a series of other classical deep neural networks (such as CNN). And the neural decision forest we used in this paper is following the structure of Peter's work. 
\begin{figure*}[!tb]
\centering
\includegraphics[width=\linewidth]{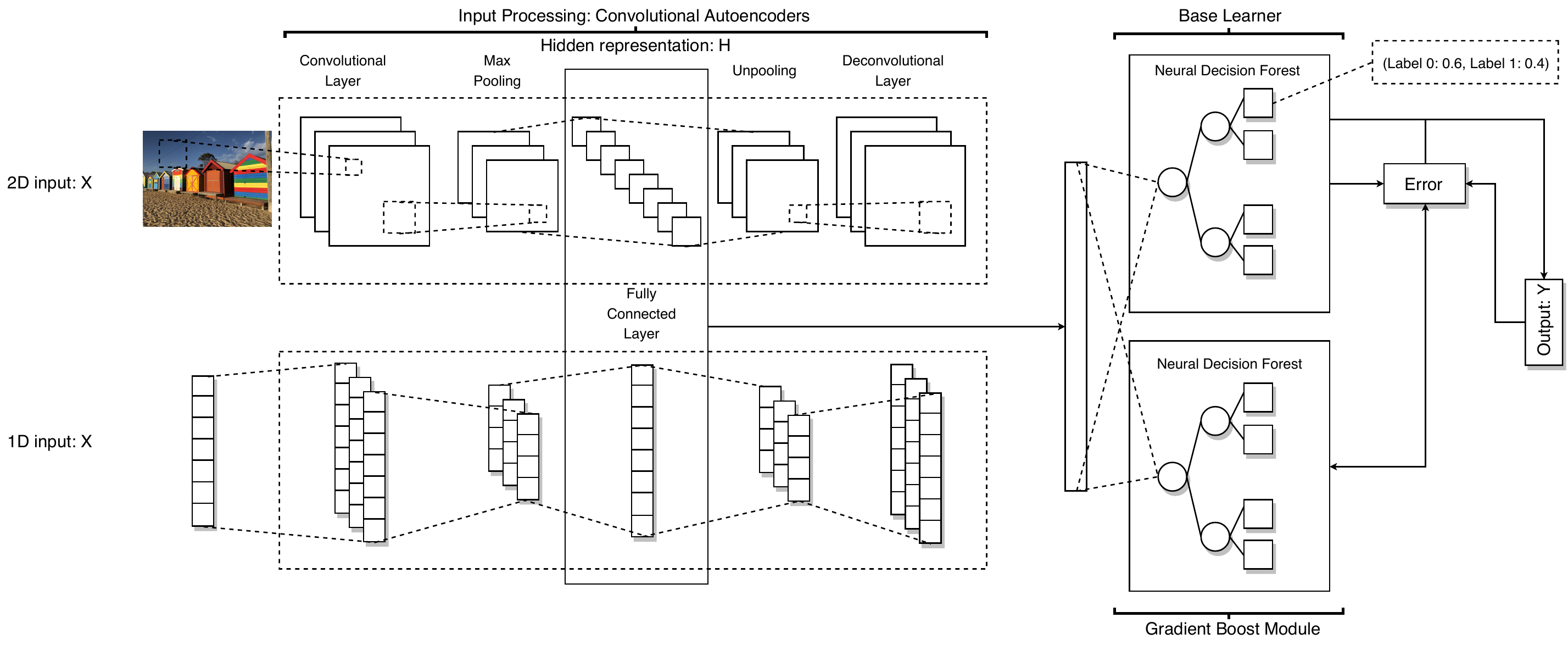}
\caption{Proposed Gradient Boosting Convolutional Autoencoder with Neural Decision Forest with one gradient boost module. The proposed model could deal with both 1D and 2D inputs.}
\label{fig:structure}
\end{figure*}
Besides improving the model itself, the idea of "boosting" is another direction for aggregate the final performance, which suggests that ensemble of models performs better than one. One of the boosting methods, gradient boost, originated from the observation by Leo Breiman \cite{breiman2001random} that boosting can be interpreted as an optimization algorithm on a suitable cost function. Based on this idea, gradient boosting machine (GBM) was subsequently developed by Friedman\cite{friedman2001greedy}, which consistently provides more accurate results than the conventional single machine learning models based on which it is designed. Friedman gives examples for adding gradient boost idea into traditional machine learning models (e.g. logistic regression), where the gradient boost decision tree(GBDT) was proposed. 
Last few years, many empirical studies showed the power of gradient boost machine in various areas and most of them focused on taking decision trees as base learners\cite{johnson2014learning}.
However, few works have applied this idea to neural networks. The first few attempts include Holger Schwenk's work \cite{schwenk2000boosting}, they employ the Adaboost idea into neural networks. Fan Zhang et. al \cite{zhang2016scene} propose to incorporate gradient boost machine with recurrent convolutional networks(RNN). To the best of our knowledge, our work is the very first to combine the gradient boost idea with neural decision forest. 

\section{The proposed Gradient Boost Convolutional Autoencoder with Neural Decision Forest}

Figure \ref{fig:structure} shows the overall work flow of proposed method with one gradient boost module, which includes four main components from left to right: input processing layer (Convolutional Autoencoder layers), fully connected layers, neural decision forest, and gradient boost module. 

The model is end-to-end trainable. Its training process includes initializing all the parameters, feeding features into the convolutional autoencoder layer $X$, and regarding the cleaned features as $X_{C}$. Then the features are processed by the hidden layer $H$, followed by fully connected layers and mapped the nodes in the tree. After that, the nodes in the decision tree layer are delivered to the nodes in $k$ decision trees where decision nodes make decisions and generate predictions by averaging the decision results. All the parameters are updated by minimizing the difference between the predicted and the real value. The bottom right part of figure \ref{fig:structure} shows the case for one gradient boost module, and the updated convolutional autoencoder and fully connected layers are shared with this gradient boost module. The gradient boost module updates their own parameters by minimizing the cost between its prediction and the error (i.e., the gap between label and neural decision forest's prediction). The final prediction is the sum of prediction of neural decision forest and the gradient boost module. The module iteratively optimizes the whole process until reaching the optimal results. We will introduce each component in details in the following subsections.

\subsection{Input Processing}
We use convolutional autoencoder as the input processing module. An autoencoder(AE) is an input processing component for producing clean feature representations in an unsupervised manner \cite{deng2014autoencoder}. It consists of two parts: the encoder and the decoder. Internally, it has a hidden layer with uses codes to represent the input. 

An auto-encoder takes $X$ (the input feature vectors) as the input and map it (with an encoder) to a hidden representation $X_{H}$, with $X_{H} = f_{E}(W_{E}X+b_{E})$, where $f_{E}(*)$ is a non-linearity function such as the sigmoid function \cite{hecht1992theory}, $W_{E}$ and $b_{E}$ are corresponding weight and bias. 
The latent representation $H$ or code is then mapped back (with a decoder) into a reconstruction $X_{C}$, $X_{C} = f_{D}(W_{D}X_{H}+b_{D})$, where $X_{C}$ is seen as a prediction of $X$, given the code $H$. Accordingly, $W_{D}$ and $b_{D}$ are weights and bias variables used in decoder part. Convolutional autoencoder has the same structures as AE but also with convolutional layers and pooling layers\cite{chen2017deep}. Both the above weights and bias variables are optimized to minimize the average reconstruction error, which is measured by the traditional squared error $\mathcal{L}_{CA}(X_{I},X_{C}) = \parallel X - X_{C} \parallel^2 $.

After the hidden layer $X_{H}$ goes through several fully connected layers, we get $X_{FC}$ as the input nodes for the neural decision forest $X_{FC} = f_{FC}(W_{FC}X_{H}+b_{FC})$, where $W_{FC}$ and $b_{FC}$ are the weights and biases used in the fully connection layers and  $f_{FC}(*)$ is the activation function\cite{hecht1992theory}. 

\subsection{Neural Decision Forest}
We introduce in this section how the input nodes $X_{FC}$ are processed through the tree layer. The neural decision forest that we use here is following the structure in Peter et al.'s work \cite{kontschieder2015deep}. 

Suppose we have a number of $T$ trees in a forest, each being a structured classifier consisting of decision (or split) nodes $D$ and prediction (or leaf) nodes $L$. The leaf nodes are the terminal nodes of the tree and each prediction node $l \in L$ holds a probability distribution $P^{t}_{l}(Y|X)$ over $Y$. Each decision node $d \in D$ in tree $t$ holds a decision function $\mathcal{D}^{t}_{d}(X;\Theta) \in [0,1]$, which stands for the probability that a sample reaches decision node $d$ and is sent to the left sub-tree. More specifically, each decision node holds the following representation:

\begin{equation}
\mathcal{D}_{d}(X;\Theta) = \sigma(f_{d}(X_{FC})) 
\end{equation}

where $\sigma(x) = (1+e^{-x})^{-1}$ is the sigmoid function, and $f_{d}(X_{FC})$ is the transfer function for $X_{FC}$. $f_{d}(X_{FC}) = W_{T}X_{FC}$, where $W_{T}$ is weight variable. $\Theta$ stands for the set of previous parameters used in autoencoder and fully connected layers: $\{W_{E},W_{D},W_{FC},W_{T},b_{E},b_{D},b_{FC}\}$. 

\begin{figure}
\centering
\includegraphics[width=0.8\linewidth]{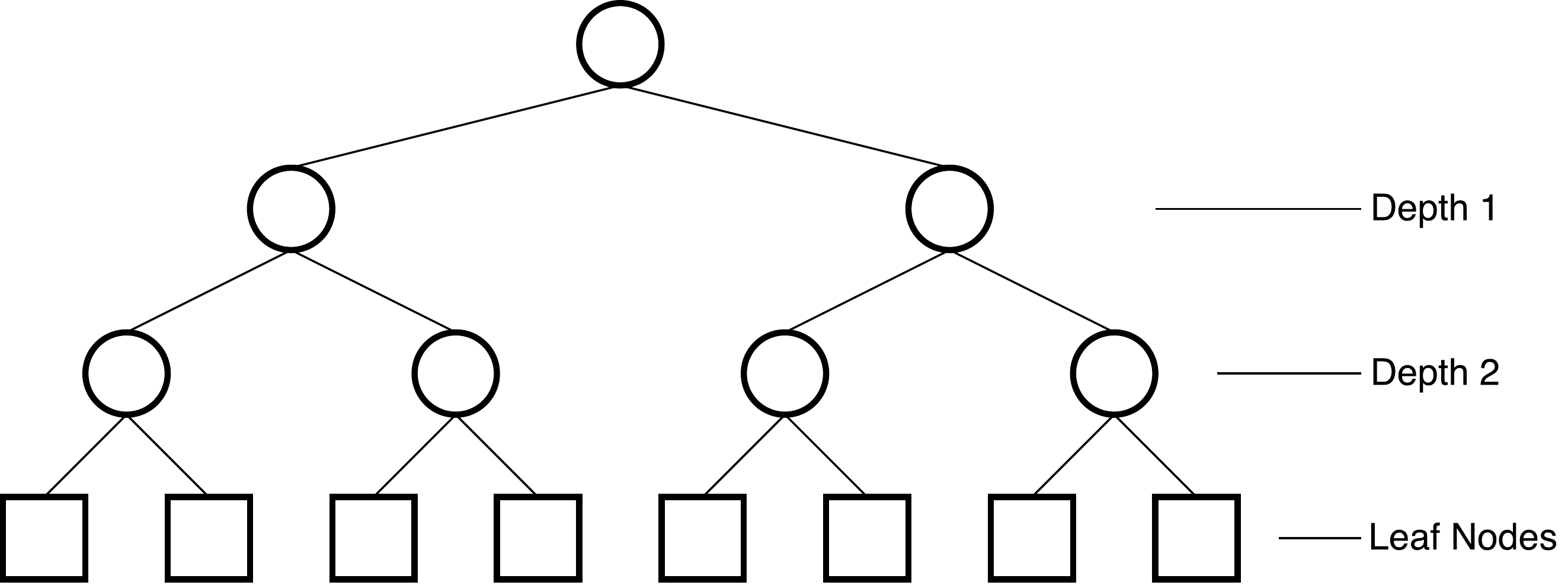}
\caption{An example of a decision tree with depth 2.}
\label{fig:tree_eg}
\end{figure}

In this paper, we suppose all the trees follow a classical binary tree structure, which means each node has two sub-trees. After defining the depth $n\_depth$ of a tree, e.g., 2, as shown in fig \ref{fig:tree_eg}, the decision nodes in depth 2 is 4 and the number of leaf nodes is 8. More generally, once the depth $n\_depth$ is defined, we can get $2^{n\_depth}$ decision nodes in depth $n\_depth$, and the number of leaf nodes, being $2^{n\_depth + 1}$. 

The probability of a sample arriving at tree $t$ and reaching in leaf $l$ is as follows:

\begin{equation}
Q_{l} = \Pi_{d \in D}\mathcal{D}_{d}(x;\Theta)^{\mathbbm{1}_{left}}\overline{\mathcal{D}_{d}}(x;\Theta)^{\mathbbm{1}_{right}},
\end{equation}

where $\overline{D_{d}}(h;\Theta) = 1 - D_{d}(h;\Theta) $, $\mathbbm{1}_{left}$ indicates the indicator function for the nodes go left. Meanwhile, the probability of this sample to be predicted as class $y$ is

\begin{equation}
\mathbb{P}_{t}[y|x,\Theta,P] = \Sigma_{l \in L} P_{l_{y}} \cdot Q_{l},
\end{equation}, 

where $P_{l_{y}}$ stands for the probability for the nodes in leaf $l$ predicted to be labeled $y$. 

For the forest of decision trees, it is an ensemble of decision trees $\mathcal{F} = \{ 1,..., T\}$ and delivers a prediction for sample $x$ by averaging the output of each tree, which can be showed from:

\begin{equation}
\mathbb{P}_{\mathcal{F}}[y|x] = \frac{1}{T}\Sigma_{t=1}^{T}\mathbb{P}_{t}[y|x]
\end{equation}

The prediction for a label, e.g., y, is as follows:
\begin{equation}
\hat{y} = \argmax_{y}\mathbb{P}_{\mathcal{F}}[y|x]
\label{eq:prediction_nrf}
\end{equation}

\subsection{Gradient Boost Module}
To put the gradient boost idea into neural networks, the samples that reach the fully connection layer $X_{FC}$ are sent to multiple destinations: one being the neural decision forest and the others gradient boost modules. 

Figure \ref{fig:structure} gives an example for only one gradient boost module. The structure of the gradient boost module is the same as the neural decision forest but has different prediction goals (predicting error) and parameters. So in the next parts, we will briefly introduce the classical gradient boost idea and then ours.


\subsubsection{Classical gradient boost algorithms}
The principle of boosting methods is to hold out optimization  in the function space\cite{friedman2001greedy}, which means the estimated function $\hat{f}(x)$ is in the additive form:

\begin{equation}
\hat{f}(x) = \hat{f^{I}}(x) = \sum^{I}_{i=0}\hat{f_{i}}(x)
\end{equation}

where $I$ is the number of iterations, $\hat{f_{0}}$ is the initial guess, $\{\hat{f_{i}}\}^{I}_{i=1}$ are the function increments (also called as "boosts"), and $\hat{f^{i}}$ is the ensemble prediction in iteration $i$.

Normally, a base learner model $h(x,\theta)$ is chose to formulate this incremental functions and the process is made in a "greedy stage-wise" way. In particular, the model sets $\hat{f_{0}}$ with $h(x,\theta_{0})$ and then choose a new function $h(x,\theta_{i})$ to be the most parallel to the negative gradient $g_{i}(x)$ along the observed data. 
\begin{equation}
g_{i}(x) = E_{y}[\frac{\partial \mathcal{L}(y,f(x))}{\partial f(x)}|x]_{f(x)=\hat{f^{i-1}}(x)}
\end{equation}
where $\mathcal{L}$ is the loss function. By setting each step with step size $\rho$, we choose the new function increment to be the most correlated with $-g_{i}(x)$. Suppose we have $N$ samples, we can handle this optimization by the classic least-squares minimization method.
\begin{equation}
(\rho_{i},\theta_{i}) = \argmin_{\rho,\theta}\sum^{N}_{j=1}[-g_{i}(x_{j})+\rho h(x_{j},\theta)]^{2}
\end{equation}

\subsubsection{Proposed gradient boost module}
In our model, the base learner $h(x,\theta)$ is an autoencoder with neural decision forest. Therefore, after learning the best initial model $f^{0} = f(x,\theta_{0})$, we get the the prediction $\hat{f^{0}} = \hat{y}$, (refer to equation \ref{eq:prediction_nrf}). Suppose we have $N$ instances and $K$ labels, then $f^{0}$ has the form: 
\begin{equation}
f^{0} = \mathbb{P}_{\mathcal{F}}[y|x] =  [\mathbb{P}_{\mathcal{F}}[y=y_{1}|x], \dots, \mathbb{P}_{\mathcal{F}}[y=y_{K}|x]]
\end{equation}
For this $K$-class problem, we solve this according to the gradient-descent boosting algorithm. The loss is defined as 
\begin{equation}\label{eq:loss}
    \mathcal{L}(y,f^{0})= - \Sigma^{K}_{k=1} y_{k} \log p_{k}(x),
\end{equation}
where $y_{k}=1(class=k)\in \{0,1\}$, and $p_{k}(x)=Pr(y_{k}=1|x)$ or equivalently,
\begin{equation}\label{eq:represent}
    p_{k}(x)=exp(\mathbb{P}(y=y_{k}))/\Sigma^{K}_{l=1}\mathbb{P}(y=y_{l})).
\end{equation}
Substituting equation \ref{eq:represent} into equation \ref{eq:loss} and taking first derivatives one has\cite{friedman2001greedy}:
\begin{align}
\nonumber G_{m} &=-[\frac{\partial L(y,f^{0}) }{\partial f^{0}}]_{f^{0}\mathbb{P}_{\mathcal{F}}[y|x]}\\
&= Y- p(x) =Y-softmax(f^{0})
\end{align}
where $G_{m}$ is the gradient "boost" that we want to use. We modify this gradient as $Error(m)=softmax(Y-f^{0})$, with the considers in the following. 

As an example, consider a binary classification problem where $K=2$ and the probability distribution estimation for sample $x$ is $(0.4,0.6)$. The sample will be predicted to be class $y_{2}$. 

We still consider the last example to illustate the finding of "boosts". Suppose $x$ actually belongs to class $y_{1}$, we hope it to take the distribution $(1,0)$ and to increase the probability that $x$ is predicted to class $y_{1}$. Matching the "boosts" $h(x,\theta_{1})$ with $(0.6,-0.6)$ to achieve the ideal distribution is unreasonable since the range of $\mathbb{P}_{\mathcal{F}}[y|x]$ is $[0,1]$. Therefore, we fit the "boosts" into a softmax version: $(\frac{e^{0.6}}{e^{0.6}+e^{-0.6}},\frac{e^{-0.6}}{e^{0.6}+e^{-0.6}})$. More specifically, for the first "boost" module, we take the next few steps: 

\begin{equation}
\left\{
\begin{aligned}
Error(1)=& softmax(Y- \mathbb{P}_{\mathcal{F}}(Y|X))= \mathbb{P}_{GB_{1}}(Y|X)\\
f^{1} =& f^{0} + \rho_{1} \cdot Error(1) \\
=&(\mathbb{P}_{\mathcal{F}}(y_{1})+ \rho_{1} \cdot \mathbb{P}_{GB_{1}}(y_{1}), \dots, \mathbb{P}_{\mathcal{F}}(y_{K}) \\ 
& + \rho_{1} \cdot \mathbb{P}_{GB_{1}}(y_{K})) \\
=& (\mathbb{P}_{1}(y_{1}),\dots, \mathbb{P}_{1}(y_{K}))\\
\overset{Scale To 1}{=}&  (\frac{\mathbb{P}_{1}(y_{1})}{\sum^{K}_{i=1}\mathbb{P}_{1}(y_{i})},\dots, \frac{\mathbb{P}_{1}(y_{K})}{\sum^{K}_{i=1}\mathbb{P}_{1}(y_{i})})\\
\hat{f}^{1} =& \argmax_{y}f^{1}\\
\end{aligned}
\right.
\end{equation}

$Error(1)$ is the prediction goal for the first gradient boost module, which is the distance between the prediction and the label. And we update the prediction $f^{1}$ by adding this distance with a step size hyper-parameter $\rho$ to the previous guess of prediction $f^{0}$. And scale the sum of each probability nodes in $f^{0}$ to 1. We will prove this step could help with the prediction in the next part. 

More generally, suppose we have $M$ gradient boost modules, and we define the prediction goal for gradient boost module $m$ as $Error(m)$. And the overall prediction in step $m$ is the sum of previous prediction and $Error(m)$. 

\begin{equation}\label{eq:kstep}
\left\{
\begin{aligned}
Error(M) =& softmax(Y-f^{M-1})= \mathbb{P}_{GB_{M}}(Y|X)\\
f^{M} =& f^{M-1} + \rho_{M} \cdot Error(M) \\
=& (\mathbb{P}_{M-1}(y_{1}) + \rho_{M} \cdot \mathbb{P}_{GB_{M}}(y_{1}),\dots, \mathbb{P}_{M}(y_{K}))\\
 & + \rho_{M} \cdot \mathbb{P}_{GB_{M}}(y_{K}) \\
 =& (\mathbb{P}_{M}(y_{1}),\dots, \mathbb{P}_{M}(y_{K})) \\
\overset{Scale To 1}{=} &  (\frac{\mathbb{P}_{M}(y_{1})}{\sum^{K}_{i=1}\mathbb{P}_{M}(y_{i})},\dots, \frac{\mathbb{P}_{M}(y_{K})}{\sum^{K}_{i=1}\mathbb{P}_{M}(y_{i})})\\
\hat{y} =&\argmax_{y}f^{M} \\
\end{aligned}
\right.
\end{equation}
After $K$ gradient boost modules, we finally get the final prediction $\hat{y}$.

\subsection{Training Process}

Following the steps in equation \ref{eq:kstep}, we have the final prediction $\hat{y}$,
\begin{equation}
    \hat{y} = \argmax_{y}f^{M} =\argmax_{y} f^{0} + \Sigma^{M}_{m=1}\rho_{m} \cdot Error(m).
\end{equation}
Since we mainly dealing with classification tasks, then we have the loss function for the prediction, which is taking the cross-entropy form\cite{lecun2015deep},  
\begin{equation}
    \mathcal{L}(y,\hat{y}) = -(y\times log(\mathbb{P}[y|x,\Theta,P]) + (1-y)log(\mathbb{P}[y|x,\Theta,P]) )
\end{equation}
where $\Theta$ stands for the parameters used in autoencoder and fully connected layer part, and $P$ stands for the parameters used in neural decision forest. 

Also, we have the loss function for the convolutional autoencoder:
\begin{equation}
    \mathcal{L}_{CA}(X_{I},X_{C}) = \parallel X - X_{C} \parallel^2
\end{equation}

Thus our optimization task is minimizing the total loss:
\begin{equation}
\mathcal{L}(x,y,\hat{y},\Theta,P) = \mathcal{L}(y,\hat{y})+\mathcal{L}_{CA}(X_{I},X_{C})
\end{equation}
which is equivalent to solving the optimal parameters:
\begin{equation}
(\Theta, P, \rho) = \argmin_{\Theta, P, \rho}(\mathcal{L}(x,y,\hat{y},\Theta,P)) 
\end{equation}

To train those parameters, we divides the learning rate by an exponentially decaying average of squared gradients using the RMSProp optimization methods\cite{ruder2016overview} and then update the parameters Sequentially. 

First, we give parameters initial random values, set down $P$ and update parameters $\Theta$ by the backpropagation from the first guess $f^{0}$,

\begin{align}
g^{\Theta}_{t} &  = \frac{\partial \mathcal{L}(\mathcal{B},\Theta,P)}{\partial \Theta} \label{eqt1}\\
G^{\Theta}_{t} &  = G^{\Theta}_{t} + g^{\Theta}_{t} \odot g^{\Theta}_{t}\\
\Theta_{t} &= \Theta_{t-1} - \frac{\eta^{\Theta}}{\sqrt{G^{\Theta}_{t}+ \epsilon}} \odot g^{\Theta}_{t} \label{eqt2}
\end{align}

where $g^{\Theta}_{t}$ is the gradient, the $\eta^{\Theta}$ and $\epsilon$ are learning rates, and $\mathcal{B}$ is a random subset of training data set. The symbol $\odot$ means the dot product. 

Then we learn the leaf node distribution probability $P$ for the first guess $f^{0}$, we, again, set down $\Theta$ and follow the RProp process but add a softmax function to the updated parameters. 

\begin{align}
g_{t}^{P} &= \frac{\partial \mathcal{L}(\mathcal{B},\Theta,P)}{\partial P} \label{eq1}\\
G_{t}^{P} &= G_{t}^{P} + g_{t}^{P} \odot g_{t}^{P} \label{eq2}\\
P_{t} &= P_{t-1} - \frac{\eta^{P}}{\sqrt{G_{t}^{P}+ \epsilon}} \odot g_{t}^{P} \label{eq3}\\
P_{t} &= softmax(P_{t}) \label{eq4} 
\end{align}
Similarly, $g^{\Theta}_{P}$ is the gradient, the $\eta^{P}$ and $\epsilon$ are learning rates. And normally the value of $\eta$ is chose as 0.001\cite{ ruder2016overview}. The details about calculating the derivative of loss function for each parameter could refer to paper \cite{kontschieder2015deep}.

Then for the gradient boost module $m$, the loss function is changed to $\mathcal{L}(x,y,f^{m},\Theta_{GB},P_{GB})$. The parameters $\Theta_{GB}$ and $P_{GB}$ for each module update in a similar way as in equation 20 to 26. As for the hyperparameter $\rho$ of the learning size for each step, normally it is solved by equation 8 but here we chose this learning size for each gradient module as a constant and we will learn this value in the next section. 

\begin{figure}[!tb]
\centering
\includegraphics[width=0.8\linewidth]{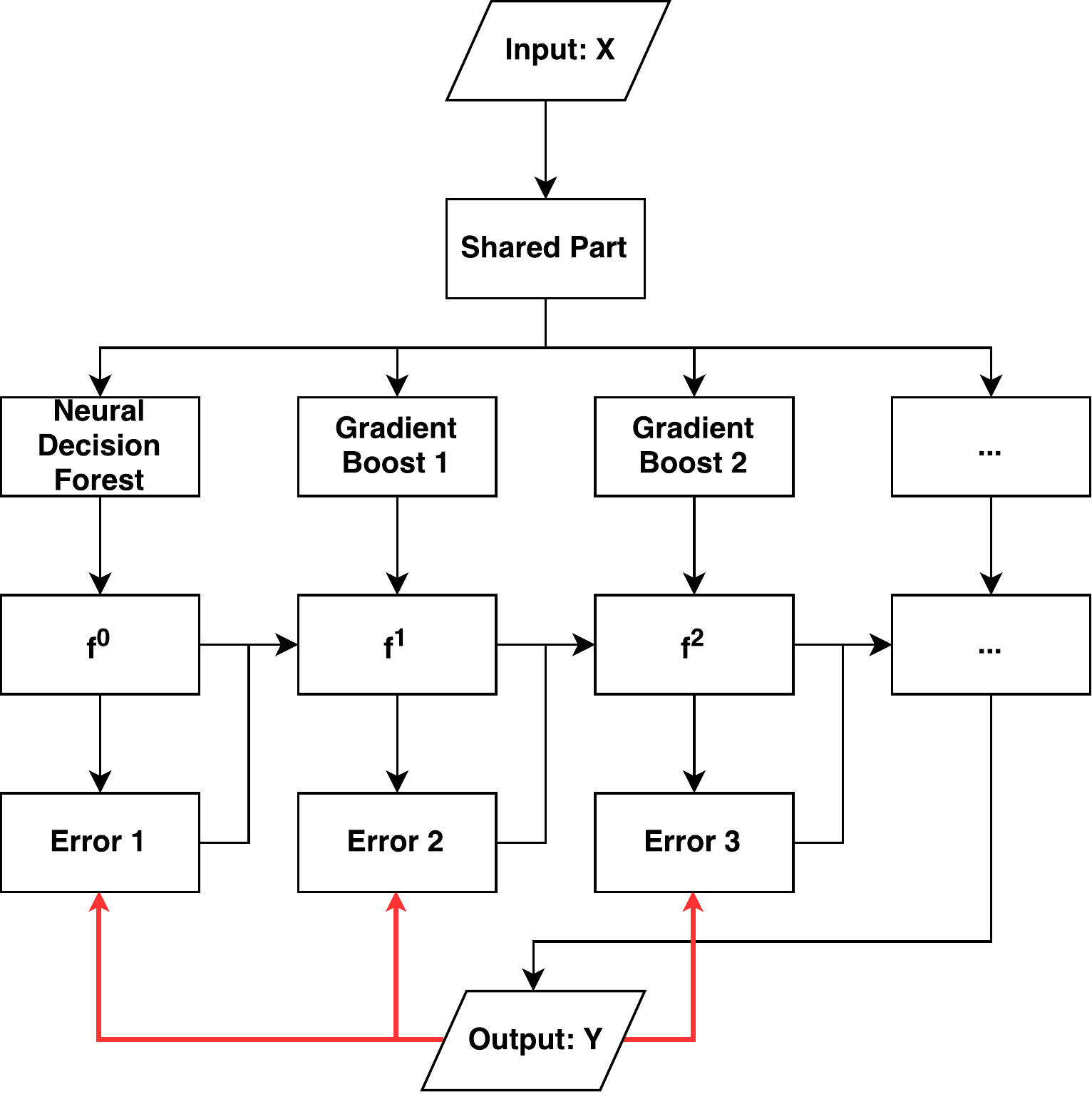}
\caption{\textbf{Flowchart for training process with multiple gradient boost modules}. The shared part is our input processing module which contains autoencoder and fully connected layers. First, it initializes all the parameters and the input data go with the black lines. Neural decision forest is our base learner, which takes the data flow and makes predictions. The label Y will give a feedback loss function to the prediction, and parameters in neural DF and shared part will be updated accordingly. The prediction and label Y together give feedback to Error 1 (through the first red line) for the gradient boost module 1 to update its parameters. Similarly, gradient boost module 2 updates its parameters according to the feedback from Error 2. For more gradient modules, they all update their parameters from previous prediction and errors.}
\label{fig:training}
\end{figure}

The training process for multiple gradient modules is shown in figure \ref{fig:training}, where the black lines show the data flow and the red lines stand for the backpropagation process.

Below shows the pseudo-code for the training and predicting process in our algorithm. 
\begin{algorithm}
\footnotesize
        \caption{Training Process}
        \begin{algorithmic}[1] 
            \Require Feature vector $X$, labels $Y$, number of epoch $n\_epoch$, number of trees $n\_tree$, depth for a tree $n\_depth$, number of gradient boost module $K$. 
            \State Generalizing the structure for the forest with $n\_tree$ and $n\_depth$
            \State Initialize parameters $\Theta, P, P_{GB}$ randomly
            \State Randomly shuffle the data sets
            \For{i \textbf{in} 1:$n\_epoch$}
                \State break data sets with $n\_batch$ of $batch\_size$ piece of data sets
                \For{j \textbf{in} 1:$n\_batch$}
                    \State Update $\Theta$ by \eqref{eqt1} to \eqref{eqt2}
                    \State Update $P$ by \eqref{eq1} to \eqref{eq4}
                    \For{k \textbf{in} 1:K}
                        \State Update $\Theta^{k}_{GB}$,$P^{k}_{GB}$
                    \EndFor
                \EndFor
            \EndFor
        \end{algorithmic}
\end{algorithm}
\subsection{Computational Complexity}
Our proposed process mainly contains the convolutional autoencoder part, the fully connected layers, the neural decision forest, and gradient boost modules. 

For each epoch, the time complexity of the 2D convolutional autoencoder(CA) part is $\mathcal{O}(\Sigma_{l}^{2*n\_CA}M^{2}K^{2} C_{l-1}C_{l})$, where $n\_{CA}$ is the number of layers from input layer to hidden layer, $M$ is the length of the output feature map, $K$ is the size of kernel, $l$ is the layer index, and $C_{l-1}$ is the number of channels in a convolutional autoencoder layer $l-1$. For 1d, $M^{2}K^{2}$ is changed to $MK$.

For the fully connected layers part, the time complexity is $\mathcal{O}(\Sigma_{l}^{n\_FC}n_{l-1}^{FC}n_{l}^{FC})$, where $n\_FC$ is the number of layers for fully connected layers, and similarly, $n_{l}^{FC}$ is the number of nodes in fully connected layer $l$.

For the neural decision forest part, the time complexity is $\mathcal{O}(n_{n\_FC}\times n\_tree \times n\_leafnodes)$, where $n_{n\_FC}$ is the number of output nodes from fully connected layers, $n\_tree$ is the number of trees in a forest, and $n\_leafnodes$ is the number of leafnodes in a tree. 

For gradient boost modules, they hold same time complexity as neural decision forest and this complexity will increase monotonously with the number of boost modules. 
For the whole training process, the whole time complexity should multiply $n\_epoch$ and $n\_batch$, which are the number of epochs and number of batch of the dataset.

\begin{algorithm}
\footnotesize
        \caption{Prediction with proposed approach(one GB module)}
        \begin{algorithmic}[1] 
            \Require Feature vector of a review $X$
            \Ensure Label for the review (fake or not) y
            \State Get hidden representation for $X$ by an encoder:
            \State $X_{H} = f_{E}(W_{E}X+b_{E})$
            \State Get tree input layer from fully connected layers:
            \State $X_{FC} = f_{FC}(W_{FC}X_{H}+b_{FC})$
            \For{t \textbf{in} 1:T}
                \State $\mathbb{P}_{t}[Y|X_{FC},\Theta,P] =  \Sigma_{l \in L} P_{l_{y}} \cdot Q_{l}$
            \EndFor
            \State $\mathbb{P}_{F}[Y|X]=\frac{1}{T}\Sigma^{T}_{t=1}\mathbb{P}_{t}[Y|X_{FC}$
            \State $Error(1)=\mathbb{P}_{GB}[Y|X]$
            \State $\hat{y}=\argmax_{y}(\mathbb{P}_{F}[Y|X]+Error(1))$
            \State \Return{y}
        \end{algorithmic}
\end{algorithm}

\section{Experiments}


In this section, we report our evaluations of classification tasks over various datasets and compared our method with several baseline works. And in the end, we do the parameter tuning to fully understand and analyze the properties of our model.

\subsection{Dataset description}
For testing our model, we valid it over different kinds of datasets, which include datasets listed in UCI machine learning repository and a public image recognition dataset. The details of datasets are listed in table \ref{tab:dataset}.

\begin{table*}[h]
\centering
\caption{Dataset Description}
\label{tab:dataset}
\footnotesize
\begin{tabular}{lllll}
  \toprule
  \textbf{Dataset Name} & \textbf{\#Classes} & \textbf{\#Instances} & \textbf{\#Attributes} & \textbf{Attribute Characteristics} \\ \midrule
  Iris & 3 & 150 & 4 & Real\\
  Wine & 3 & 178 & 13 & Integer, Real\\
  Breast Cancer Wisconsin (Diagnostic) & 2 & 699 & 9 & Real \\
  Epileptic Seizure Recognition Data Set & 5 & 11500 & 178 & Integer, Real \\
  Fashion MNIST & 10 & 70000 & 784 & Integer(Pixel) \\
  \bottomrule
\end{tabular}
\end{table*}

The UCI Machine Learning Repository\footnote{UCI Machine Learning Repository \url{https://archive.ics.uci.edu/ml/}} has been widely used as a primary source of machine learning data. And the first three datasets are the most popular datasets in UCI, and they are of small training size. Iris dataset contains 3 classes of 50 instances each, and the four attributes are describing the length and width information of sepal and petal. Wine dataset is using chemical analysis determine the origin of wines, the analysis determined the quantities of 13 constituents found in each of the three types of wines. The features of Breast Cancer Wisconsin dataset are computed from a digitized image of a fine needle aspirate (FNA) of a breast mass, and nine real-valued features are computed for each cell nucleus, each instance has a label of malignant or benign. 

The Epileptic Seizure Recognition Dataset\cite{andrzejak2001indications} is a pre-processed and re-structured version of a very commonly used dataset featuring epileptic seizure detection. The original dataset has total 500 individuals with each has 4097 data points for 23.5 seconds. Each data point is the value of the EEG recording at a different point in time. The reshaped one shuffled every 4097 data points into 23 chunks, each chunk contains 178 data points for 1 second. Thus there are total 11500 pieces of information(row), each information contains 178 data points for 1 second(column). Each instance will be labeled as one of the followings: "eyes open", "eyes closed", "they identify where the region of the tumor was in the brain and recording the EEG activity from the healthy brain area",  "they recorder the EEG from the area where the tumor was located", or "recording of seizure activity". 

\begin{figure}[!tb]
\centering
\includegraphics[width=0.9\linewidth]{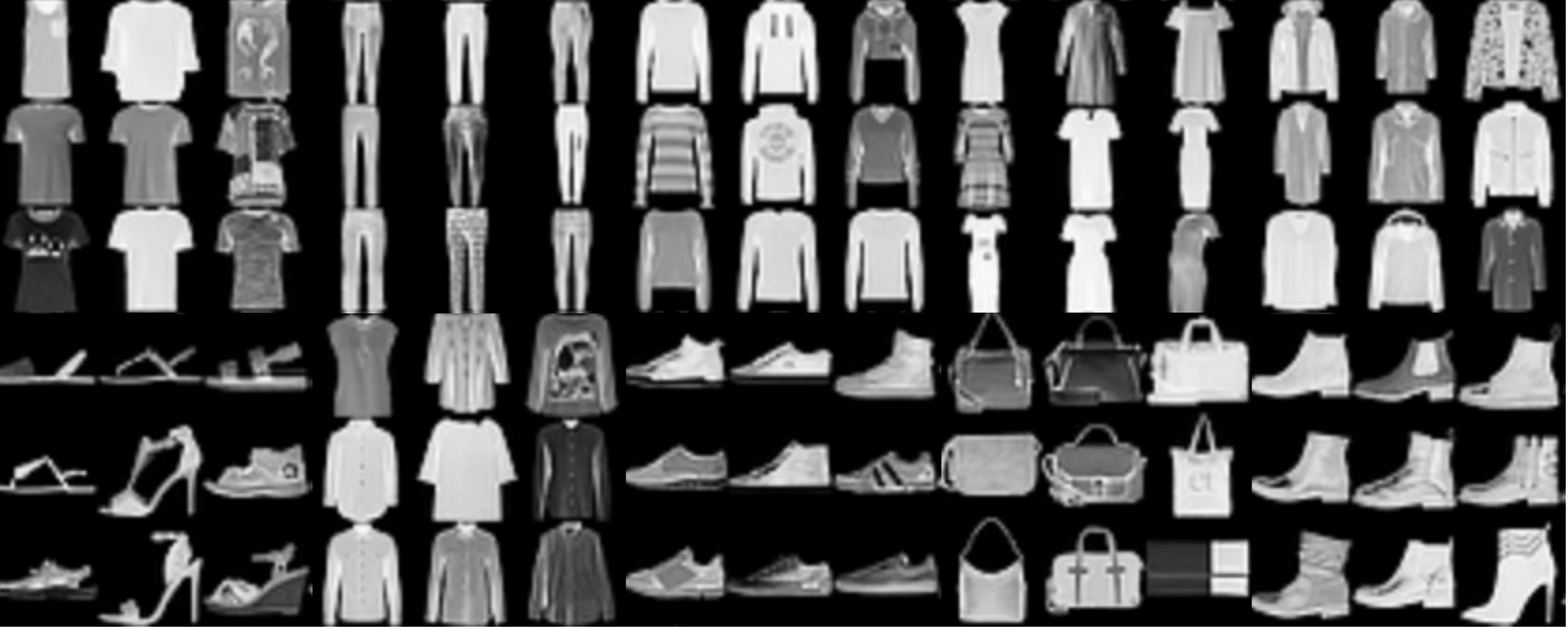}
\caption{Example for fashion MNIST. Each class is represented by nine cases.}
\label{fig:fashion}
\end{figure}

Fashion-MNIST\cite{xiao2017fashion} is a dataset of Zalando's article images. This dataset consists of a training set of 60,000 samples and a test set of 10,000 samples. Each sample is associated with a label from a total of 10 classes: T-shirt/top(label 0), trouser(1), pullover(2), dress(3), coat(4), sandal(5), shirt(6), sneaker(7), bag(8), ankle boot(9). Each image is 28 pixels in height and 28 pixels in width. Each pixel has an integer pixel-value between 0 and 255 associated with it, indicating the lightness or darkness of that pixel. An example of the images is shown in figure \ref{fig:fashion}.

\subsection{Settings}
In this part, we will introduce the parameter settings for different datasets and present the details of the comparison methods as well as the prediction evaluation metrics.

\subsubsection{Parameter Settings}
The Neural network is proven to be good at dealing with large-scale datasets\cite{zhou2017deep} and usually, the number of parameters in deep learning models is quite large. However, when used on small datasets, a deep neural networks model might suffer over-fitting since the number of parameters might be even larger than the number of instances. Thus for the first three datasets (Iris, Wine and Breast Cancer Wisconsin), we control the number of parameters in a reasonable scale to avoid the over-fitting, as detailed below.

We randomly separated those datasets into 80\% training data and 20\% test data (if training/test data not predefined) with the samples with missing values discarded. For the model settings, we used two convolutional autoencoders and afterwards one fully connected layer. Also, we configured one step for the gradient boost module and a depth of 3 for both gradient boost module and neural decision forest. The batch size was around ten percent of the whole dataset. The number of nodes of each neural network layer was decided according to the number of features and not greater than 4 times of the number of features. All weights and bias were initialized with randomly distributed values. And each experimental setup was iterated 200 times.

For the other two datasets, we take similar strategies. Where for Epileptic Seizure Recognition Data Set, we randomly separate the dataset into 8000 training samples and 3500 testing samples and then transform the label into the one-hot format. For Fashion MNIST, which contains 60000 training samples and 10000 test samples, we keep its shape in height of 28 pixels and in width of 28 pixels, and we also transform the label into the one-hot format.
When using comparison methods without CNN module, we tile each image matrix into a vector with 784 nodes. The learning for the parameters is listed in section 4.4. 

\subsubsection{Comparison Methods}
We compare three baseline methods most related to our proposed models, namely gradient boost decision trees(GBDT), random forest, and convolutional neural networks(CNN). Their detailed settings are listed below. 
\begin{itemize}
    \item Gradient Boost Decision Trees(GBDT)\cite{friedman2001greedy}: is the boost version of decision trees, which is following the Friedman's design. The number of estimators is 100, the max depth of a tree is 10, and the loss function is deviance. 
    \item Random Forest\cite{ho1995random}: is known as the embedding version of decision trees. Here we choose the number of estimators(trees) as 100, the max depth as 100 and the criterion as entropy. 
    \item Convolutional Neural Networks(CNN)\cite{lecun2015deep}: is quite popular these years. The convolution emulates the response of an individual neuron to visual stimuli, and each convolutional neuron processes data only for its receptive field. The settings are two convolutional neural with max pooling. 
\end{itemize}

\begin{table*}[!htb]
\centering
\caption{Comparison with Baseline Methods on Epileptic Dataset}
\label{tab:result}
\begin{tabular}{lcccccccc}
  \toprule
   & \multicolumn{4}{c}{\textbf{Epileptic Dataset}} & \multicolumn{4}{c}{\textbf{Fashion MNIST}} \\ \midrule
  \textbf{Learner} & \textbf{Accuracy} & \textbf{Precision} & \textbf{Recall} & \textbf{F1-score} & \textbf{Accuracy} & \textbf{Precision} & \textbf{Recall} & \textbf{F1-score} \\ \hline
  GBDT & 0.4961 & 0.4937 & 0.4992 & 0.4847 & 0.8837 & 0.8820 & 0.8840 & 0.8824 \\
  Random Forest & 0.5684 & 0.5695 & 0.5668 & 0.5679 & 0.8790 & 0.8764 & 0.8787 & 0.8764\\
  CNN & 0.7531 & 0.7496 & \textbf{0.7673} & 0.7510 & 0.8994 & 0.8994 & 0.8987 & 0.8985 \\ 
  Peter et al. & 0.7493 & 0.7456 & 0.7473 & 0.7462 & 0.8564 & 0.8566 & 0.8589 & 0.8564 \\
  Zhang et al. & 0.7485 & 0.7468 & 0.7481 & 0.7464 & 0.9164 & 0.9163 & 0.9173 & 0.9149 \\
  Li et al.    & 0.7232 & 0.7241 & 0.7262 & 0.7244 & 0.8860 & 0.8860 & 0.8912 & 0.8870 \\
  Dennis et al. & 0.7479 & 0.7464 & 0.7461 & 0.7460 & 0.9159 & 0.9159 & 0.9178 & 0.9159 \\
  \midrule
  \textbf{CAN} & 0.7570 & 0.7546 & 0.7549 & 0.7544 & 0.9150 & 0.9072 & 0.9083 & 0.9075 \\
  \textbf{GrCAN} & \textbf{0.7650} & \textbf{0.7611} & 0.7611 & \textbf{0.7602} & \textbf{0.9201} & \textbf{0.9201} & \textbf{0.9203} & \textbf{0.9201} \\
  \bottomrule
\end{tabular}
\label{tab:states}
\end{table*}

Also, we compare our work with four most related state-of-arts. 
\begin{itemize}
    \item Peter et al.\cite{kontschieder2015deep}: they proposed deep neural decision forests, which is also used in our model. The original work combined deep neural decision forests with convolutional neural networks to deal with image recognition tasks. For the comparison, we also follow the original design and set around 50 trees in a forest.
    \item Zhang et al.\cite{zhang2016scene}: proposed gradient boost random convolutional neural networks, which uses the convolutional neural network as the base learner. The base learner and the "boost" are two convolutional layers followed with one fully connected layer.
    \item Li et al.\cite{li2017deep}: contains an autoencoder and a special prototype layer, where each unit of that layer stores a weight vector that resembles an encoded training input followed by a fully connected layer. 
    \item Dennis et al. \cite{hamester2015face}: designed a two-channel convolutional neural network, where a channel use traditional CNN filters and another one use convolutional autoencoders as filters. 
\end{itemize}
Besides, we also compared the proposed structure with the base learner: Convolutional Autoencoder with Neural decision forest(CAN), and then our proposed Gradient boost Convolutional Autoencoder with Neural decision forest(GrCAN).

\subsubsection{Performance metrics}
To evaluate the performance of the classification, we apply the following criteria. For a binary classification problem, the predicted value could be either true positive (TP), false positive (FP), false negative (FN), and true negative (TN). The four criteria for classification are defined as:
\begin{itemize}
    \item Accuracy: $\frac{TP+TN}{TP+TN+FP+FN}$ 
    \item Precision: $\frac{TP}{TP+FP}$
    \item Recall: $\frac{TP}{TP+FN}$
    \item F1-score:  $2\cdot \frac{precision \cdot recall}{precision + recall}$
\end{itemize}
where TP is the number of correctly predicted positive objectives, TN is the number of correctly predicted negative objectives, FP is the number of falsely predicted positive objectives and FN is the number of falsely predicted negative objectives. For a multi-classification problem, we here take the one-vs-rest strategy, which means we regard one label as positive and others are negative ones, and taking the average of the corresponding criteria for the total results.

\subsection{Evaluation}
We firstly report the comparison of our model with a series of baseline methods and state-of-arts on two larger scale datasets: Epileptic Seizure Recognition dataset and Fashion MNIST. Table \ref{tab:result} shows the comparison results. The second to fifth columns are results for Epileptic Dataset. And the last four columns are for Fashion MNIST, a ten classes classification problem. The first three rows are three baseline methods and followed with four state-of-arts, the last two are our base learner CAN and our proposed GrCAN. 

We could see that neural networks based methods normally outperform the boosting or ensemble of decision trees (GBDT and Random Forest). And models contain CNN module (Peter et al., Zhang et al., Dennis et al. and our designs) perform better than the model without CNN modules, this might result from the CNN's capability for capturing the neighborhood information especially for dealing with image dataset. And compare our base learner CAN with the boosted version GrCAN, we could see that adding gradient boost module to help with improving the accuracy of the base learner. And in general, our proposed model, which combined neural decision forest with gradient boosting ideas, performs better than most of the baseline methods. Therefore, we conclude that the proposed gradient boost neural decision forest model is competitive, if not the best, classification model for image recognition tasks. 

\begin{figure}[htbp]
\centering
\includegraphics[width=0.9\linewidth]{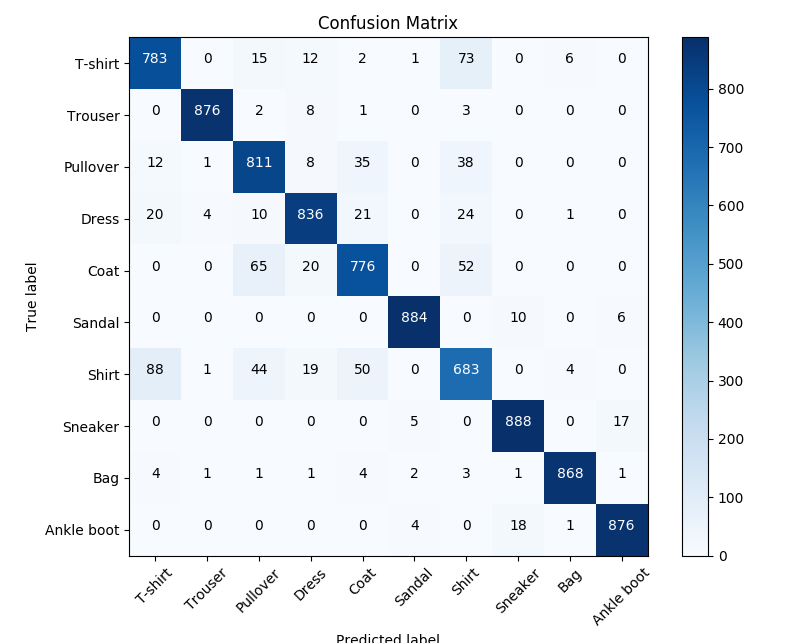}
\caption{Confusion Matrix on the prediction Fashion MNIST dataset}
\label{fig:confusion matrix}
\end{figure}

Figure \ref{fig:confusion matrix} gives a confusion matrix about the details about our prediction for Fashion-MNIST. We could see that instances that labeled as T-shirt, Pullover, Coat or Shirt are most difficult to discriminate. Which is also in our common sense that these four types of clothes sometimes look similar to each other. 

Also, we present the performance of our model on three datasets with moderate data size: Breast Cancer dataset, Iris dataset and Wine dataset. 

\begin{figure}[!tb]
\centering
\begin{minipage}[t]{4.4cm}
\includegraphics[width=4.4cm]{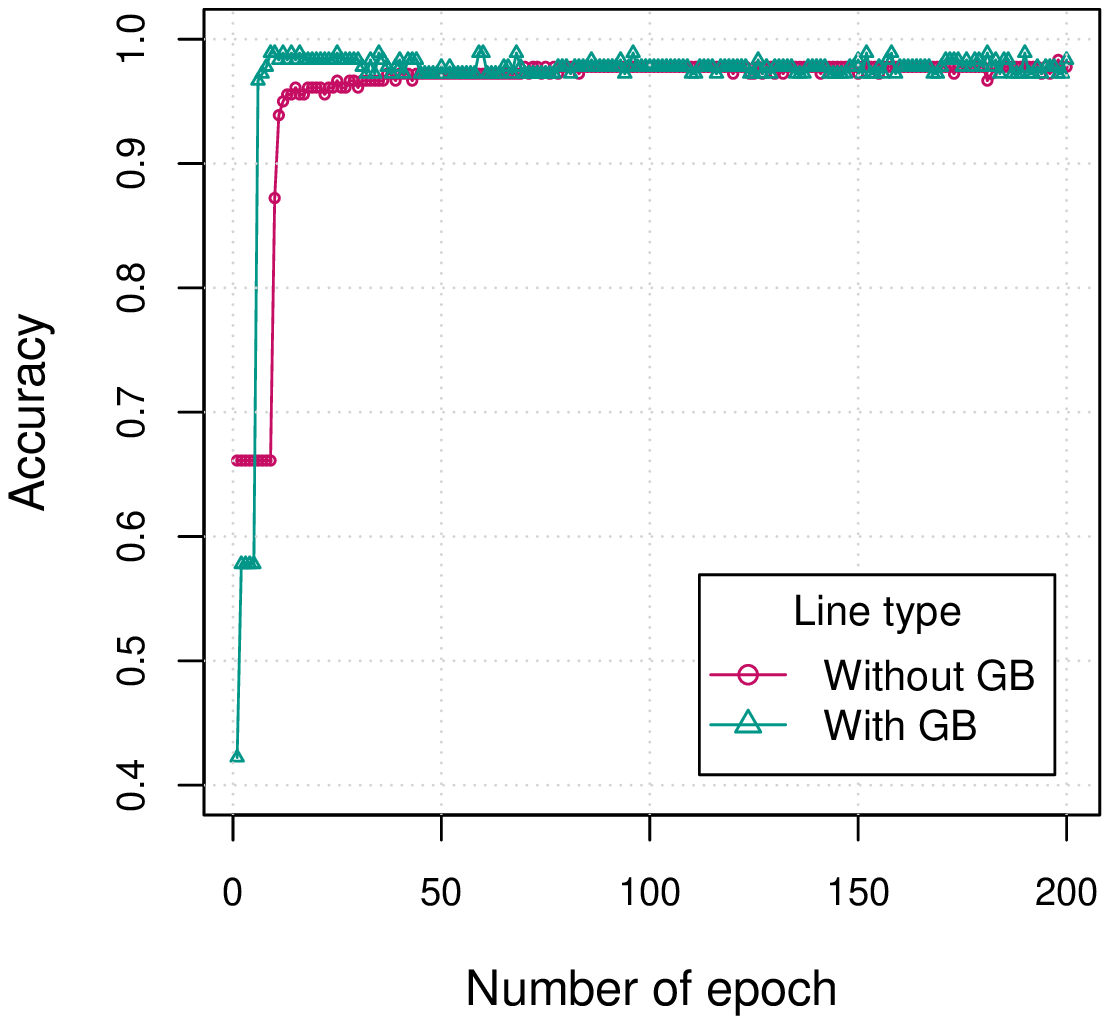}
\centering{(a)Breast Cancer: accuracy}
\end{minipage}
\begin{minipage}[t]{4.4cm}
\includegraphics[width=4.4cm]{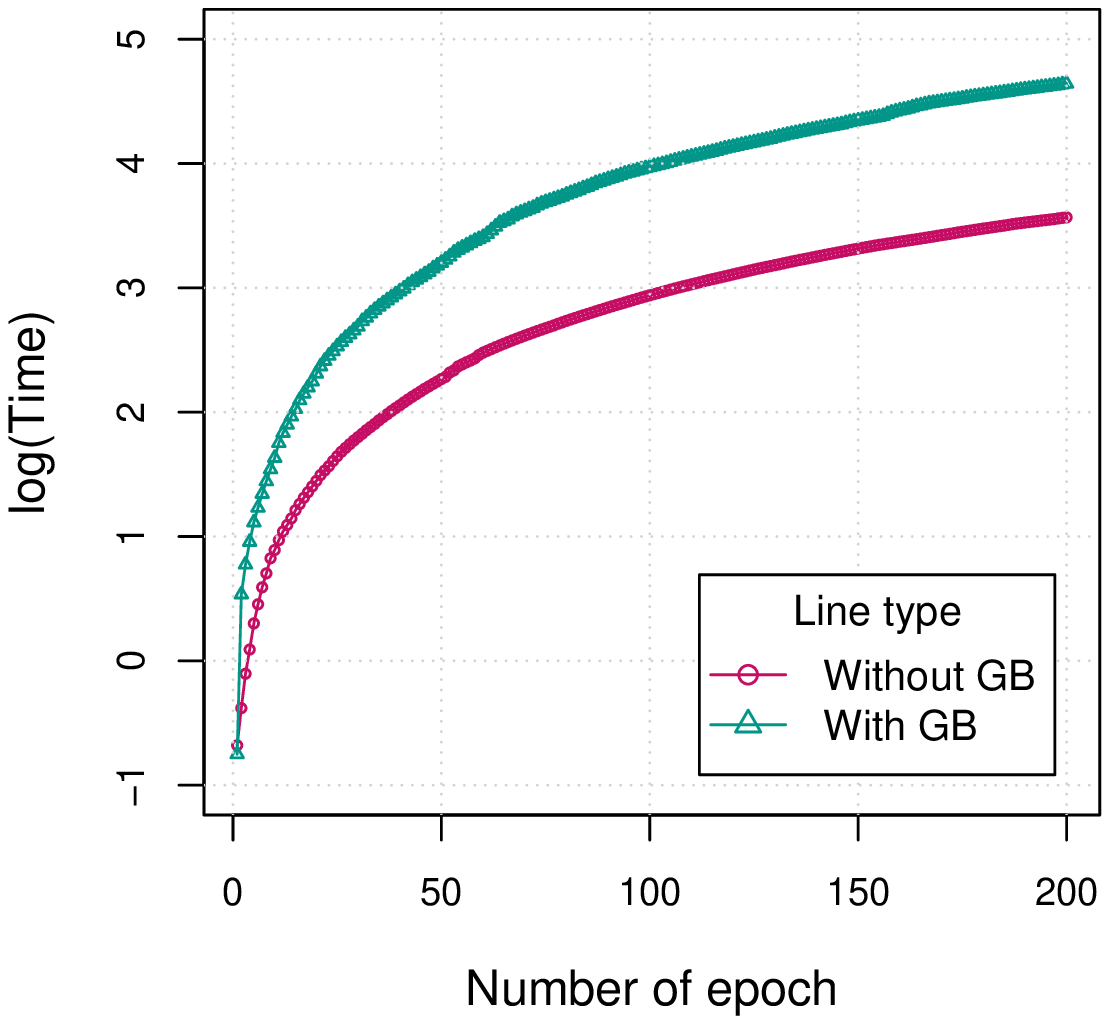}
\centering{(b)Breast Cancer: time}
\end{minipage}
\begin{minipage}[t]{4.4cm}
\includegraphics[width=4.4cm]{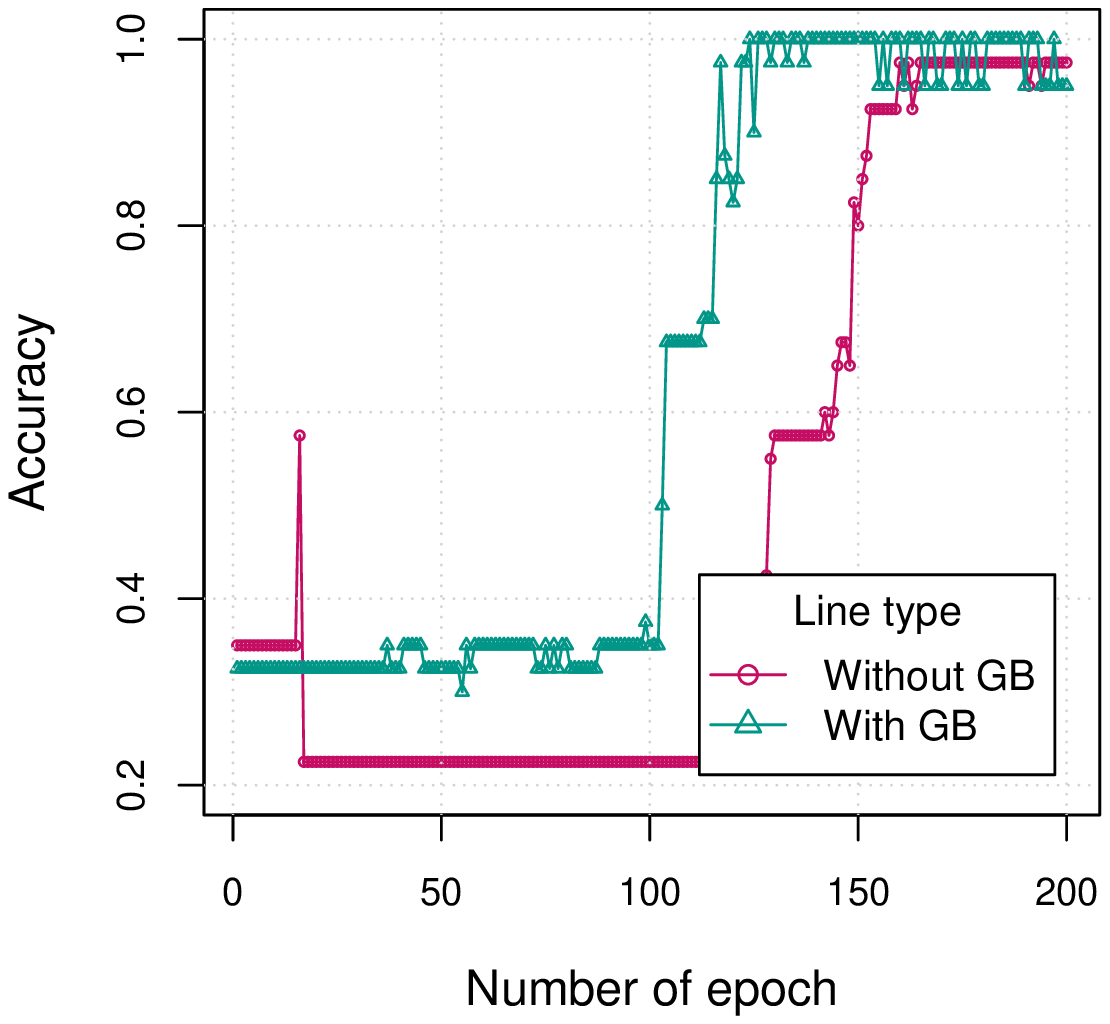}
\centering{(c)Iris: accuracy}
\end{minipage}
\begin{minipage}[t]{4.4cm}
\includegraphics[width=4.4cm]{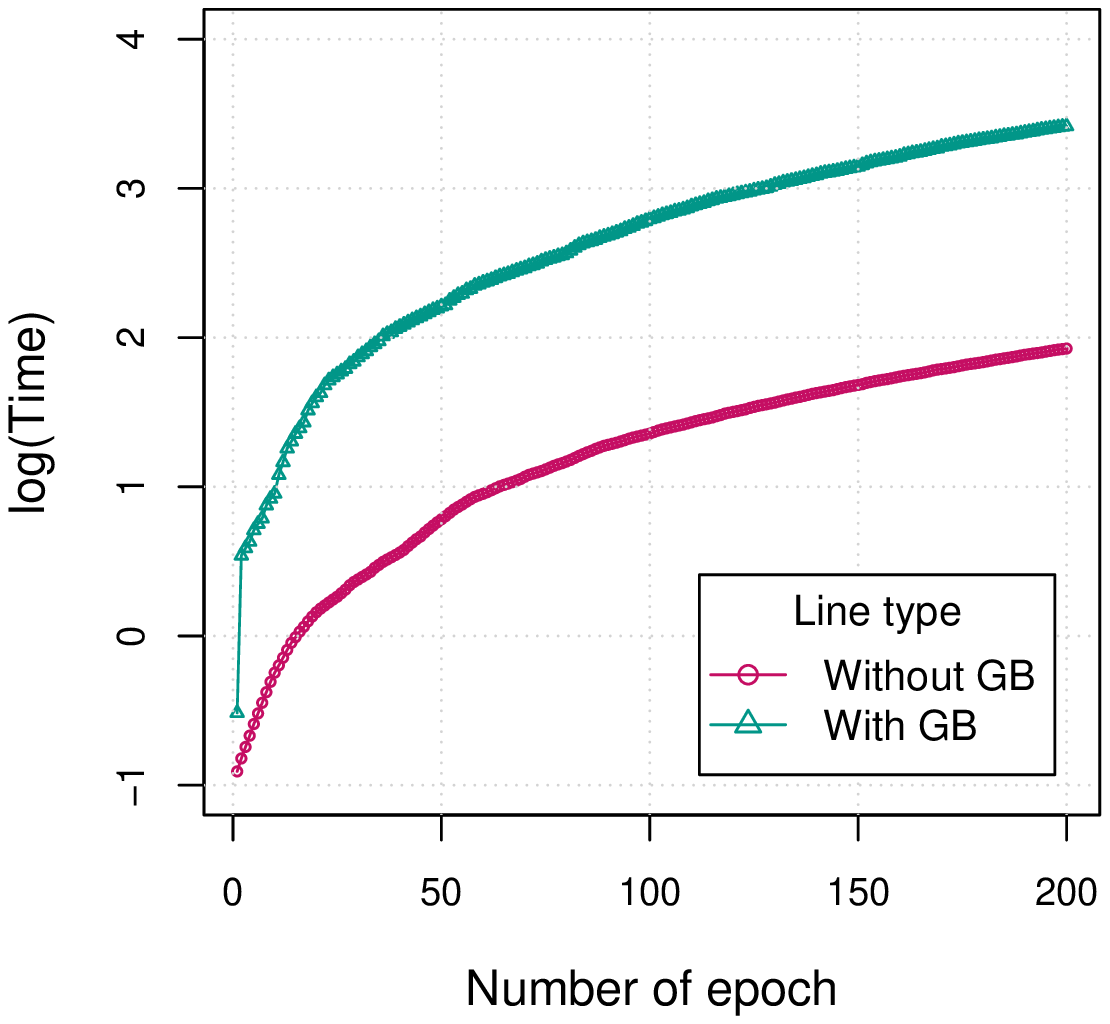}
\centering{(d)Iris: time}
\end{minipage}
\begin{minipage}[t]{4.4cm}
\includegraphics[width=4.4cm]{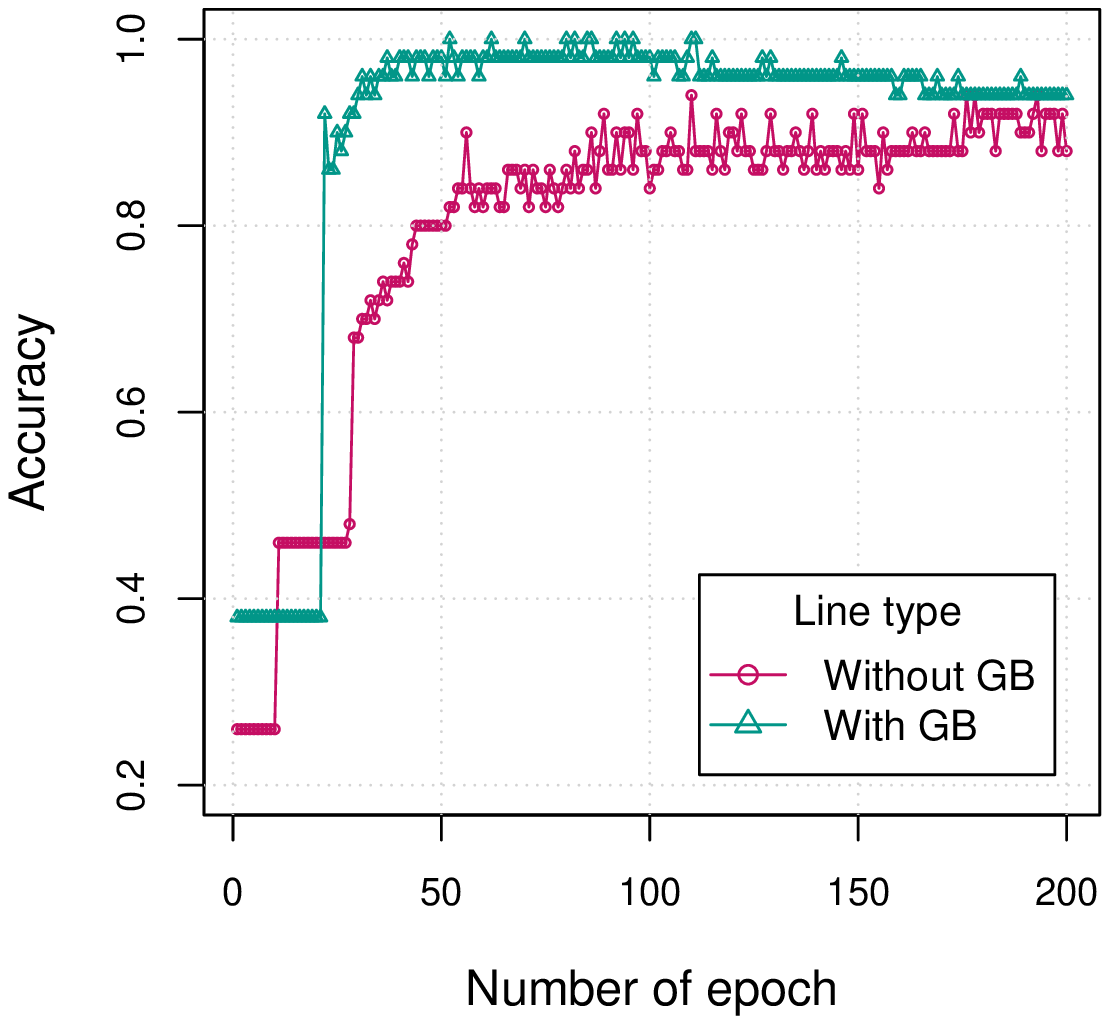}
\centering{(e)Wine: accuracy}
\end{minipage}
\begin{minipage}[t]{4.4cm}
\includegraphics[width=4.4cm]{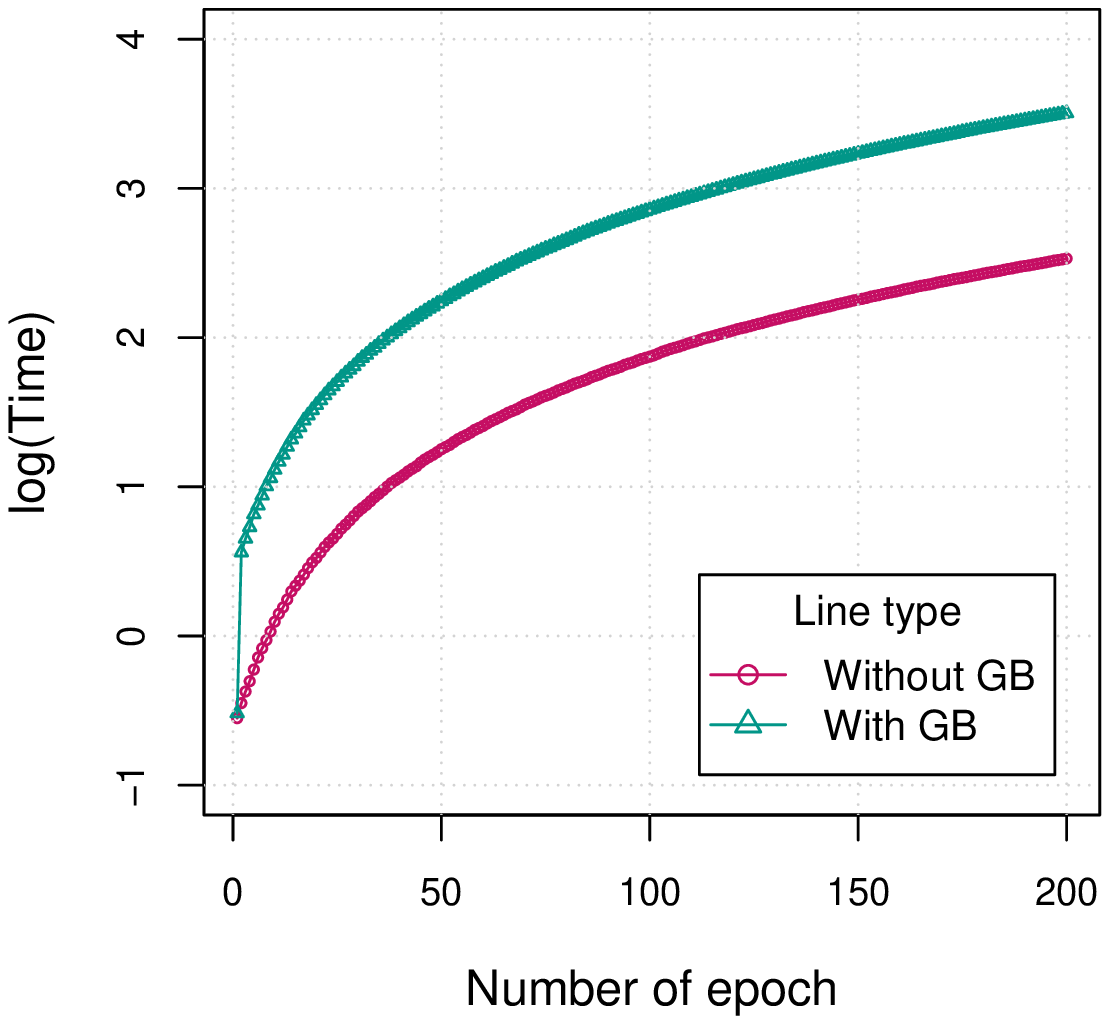}
\centering{(f)Wine: time}
\end{minipage}
\caption{Comparison of with or without gradient boost module for Breast Cancer, Iris, and Wine dataset in accuracy((a),(c),(e)) and time((b)(d)(f)). }
\label{fig:uci}
\end{figure}

Figure \ref{fig:uci} shows the accuracy and time cost for these three datasets. And we mainly compare the performance of the model with or without gradient boost modules, where the red lines stand for the only autoencoder with neural decision forest model and green lines stand for our proposed model. The time is in a log form. The results show our model achieves over 95\% accuracy over these three datasets, demonstrating that incorporating the gradient boost module improve the prediction performance. On the other hand, adding the gradient boost module increases the time cost of each iteration, but the accuracy reaches an acceptable level faster. Therefore, the overall performance is still satisfactory, and prove our model's efficiency in dealing with small size dataset.

\subsection{Parameter Tuning}
In this part, we will learn the parameters effects on our model. As mentioned above, we mainly have parameters in the base learner: the input processing part (convolutional autoencoder and fully connected layers) and neural decision forest, as well as hyper-parameters for the gradient boost. 

\begin{figure}[htbp]
\centering
\includegraphics[width=0.8\linewidth]{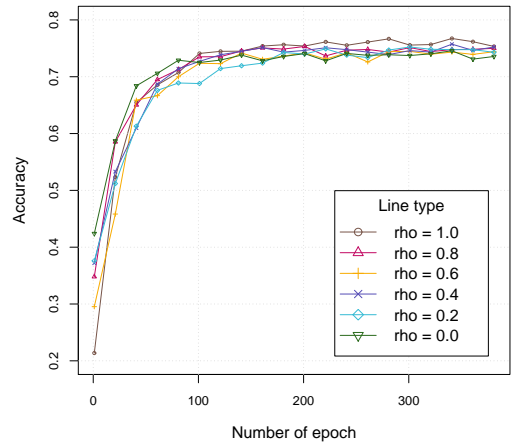}
\caption{Learning for hyper-parameter $\rho$}
\label{fig:parameter_rho}
\end{figure}

The hyper-parameters contain parameters in optimization steps and gradient boost step size $\rho$. And here we represent the learning of $\rho$ over Epileptic Dataset, which is shown in figure \ref{fig:parameter_rho}. 

We could see that bigger $\rho$ needs more iterations to convergence but is more helpful for the gradient module. Such as when $\rho = 1.0$, the accuracy in the first few iterations performs worse than the prediction with other $\rho$ values, but then the accuracy increases fast in the latter few iterations and performs the best in the end.

\begin{figure*}[!htb]
\begin{minipage}[t]{5.7cm}
\includegraphics[width=\textwidth]{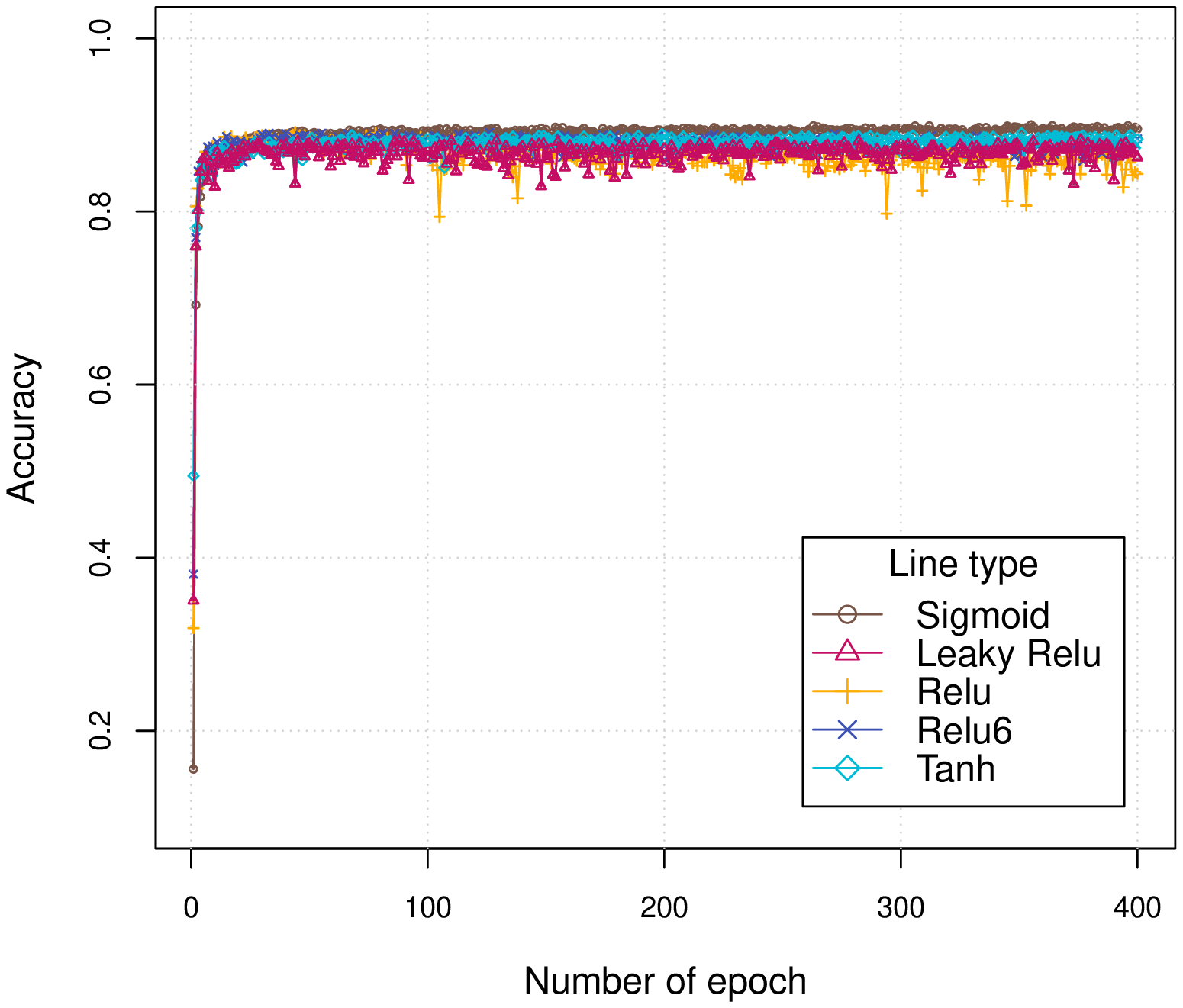}
\centering{(a)}
\centering
\end{minipage}
\begin{minipage}[t]{5.7cm}
\includegraphics[width=\textwidth]{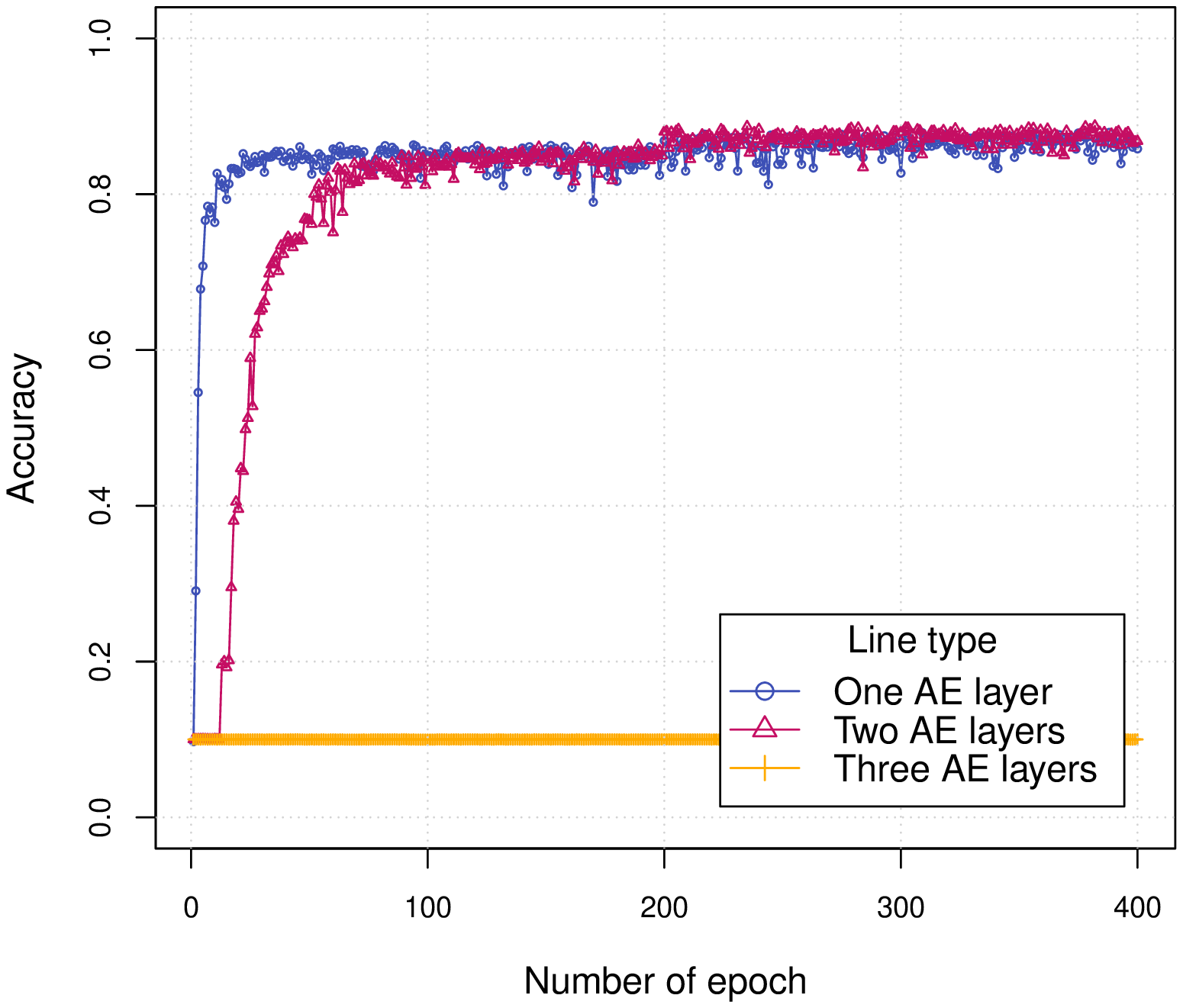}
\centering{(b)}
\centering
\end{minipage}
\begin{minipage}[t]{5.7cm}
\includegraphics[width=\textwidth]{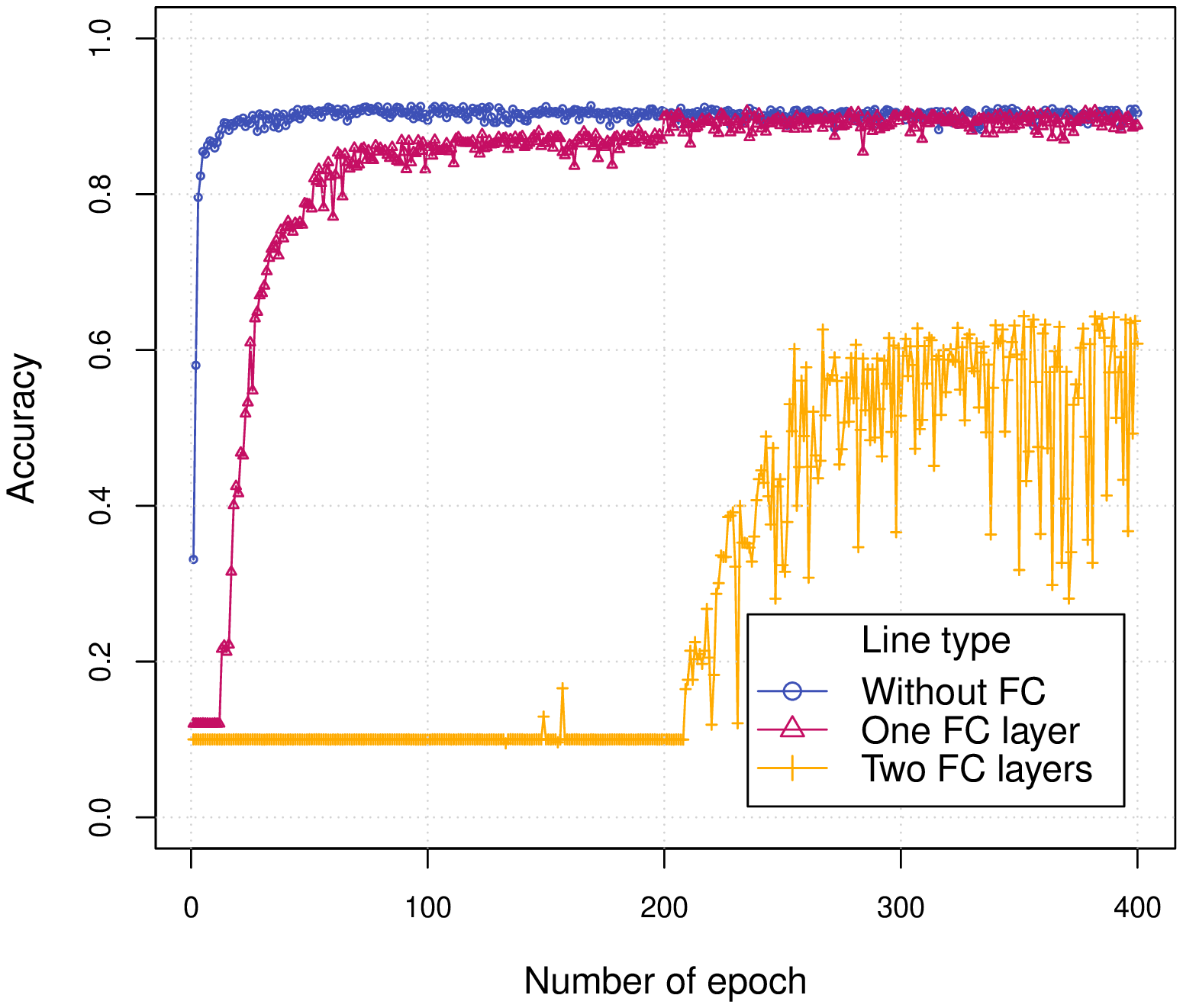}
\centering{(c)}
\centering
\end{minipage}
\begin{minipage}[t]{5.7cm}
\includegraphics[width=\textwidth]{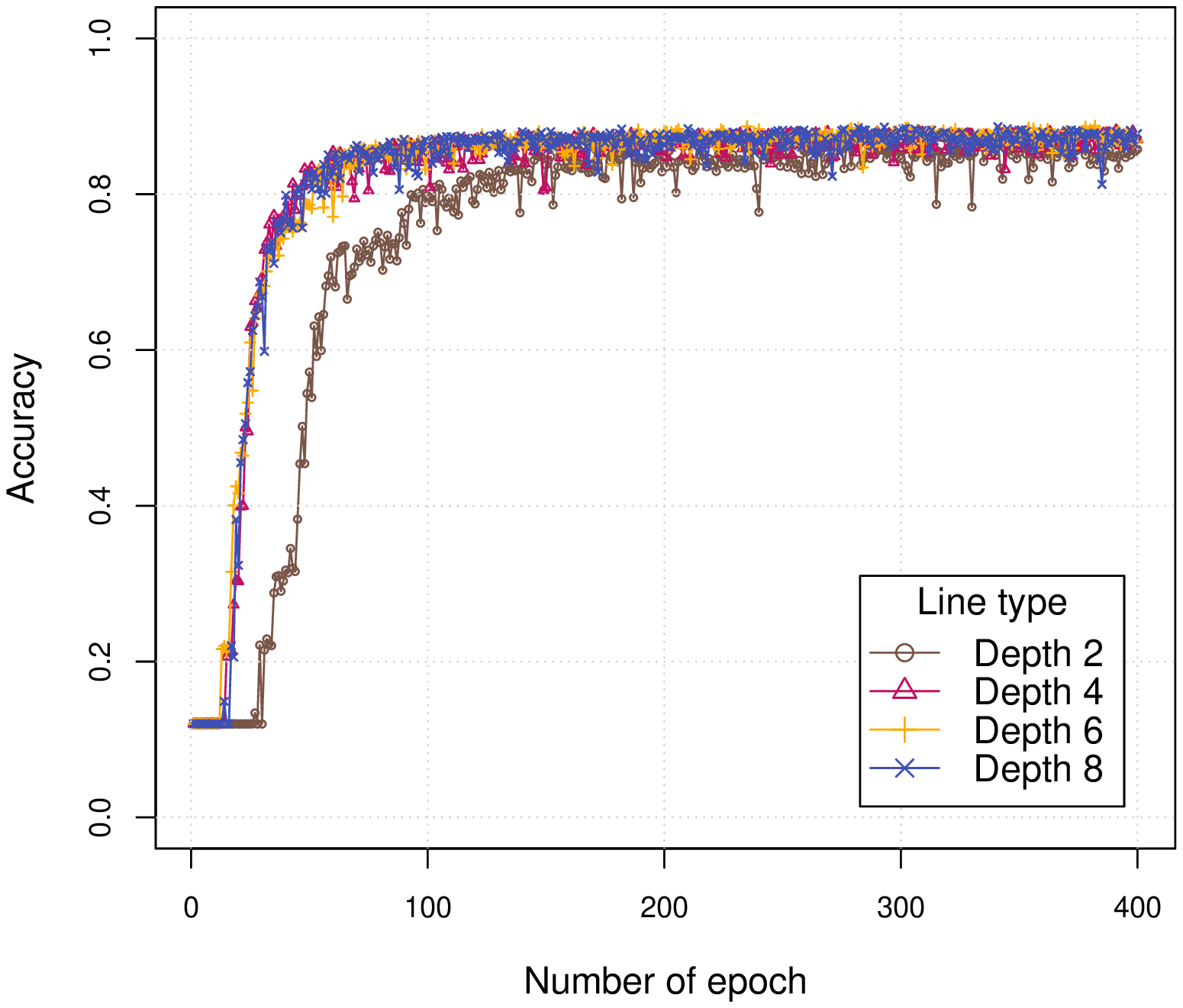}
\centering{(d)}
\centering
\end{minipage}
\begin{minipage}[t]{5.7cm}
\includegraphics[width=\textwidth]{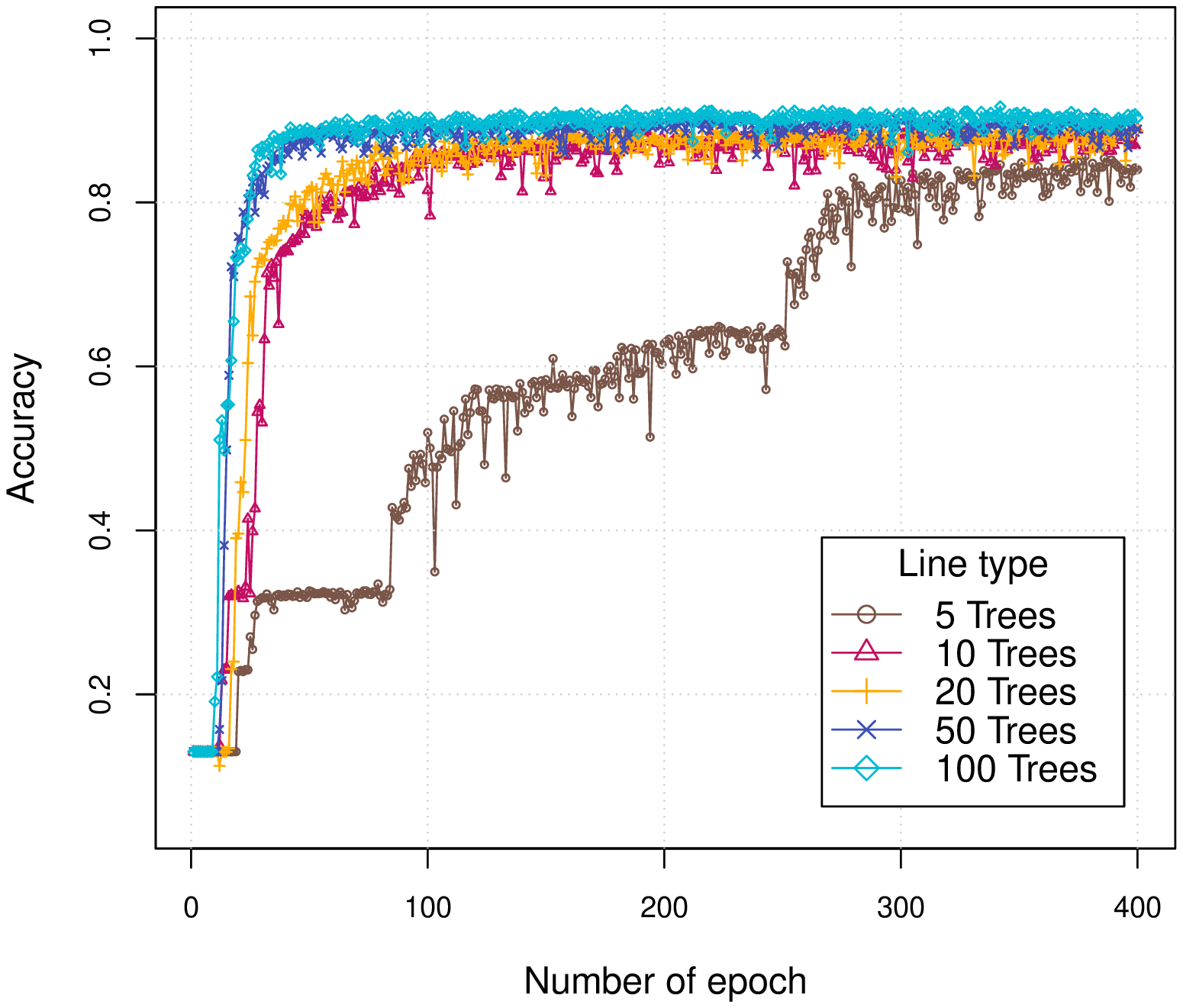}
\centering{(e)}
\centering
\end{minipage}
\begin{minipage}[t]{5.7cm}
\includegraphics[width=\textwidth]{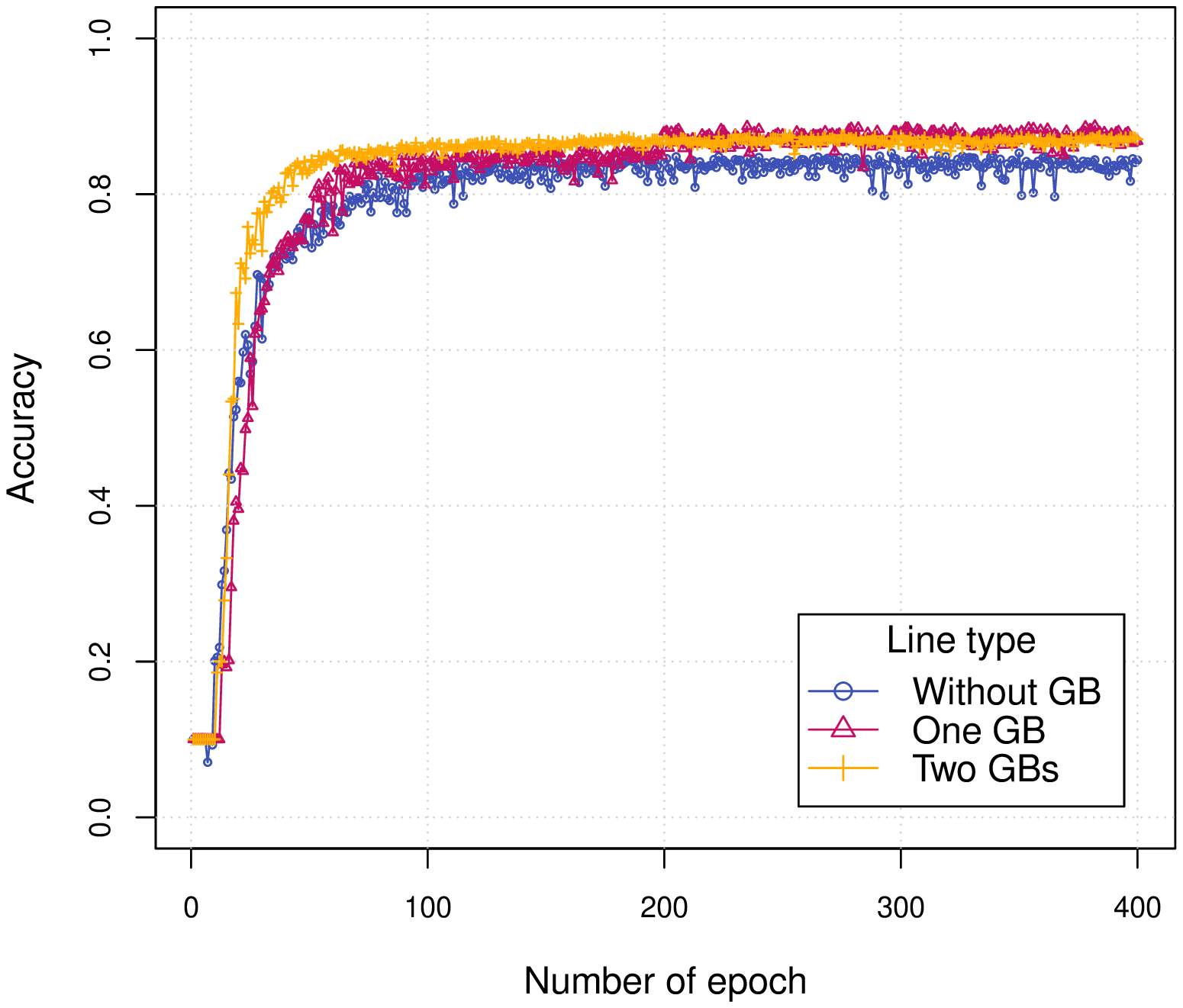}
\centering{(f)}
\centering
\end{minipage}
\begin{minipage}[t]{5.7cm}
\includegraphics[width=\textwidth]{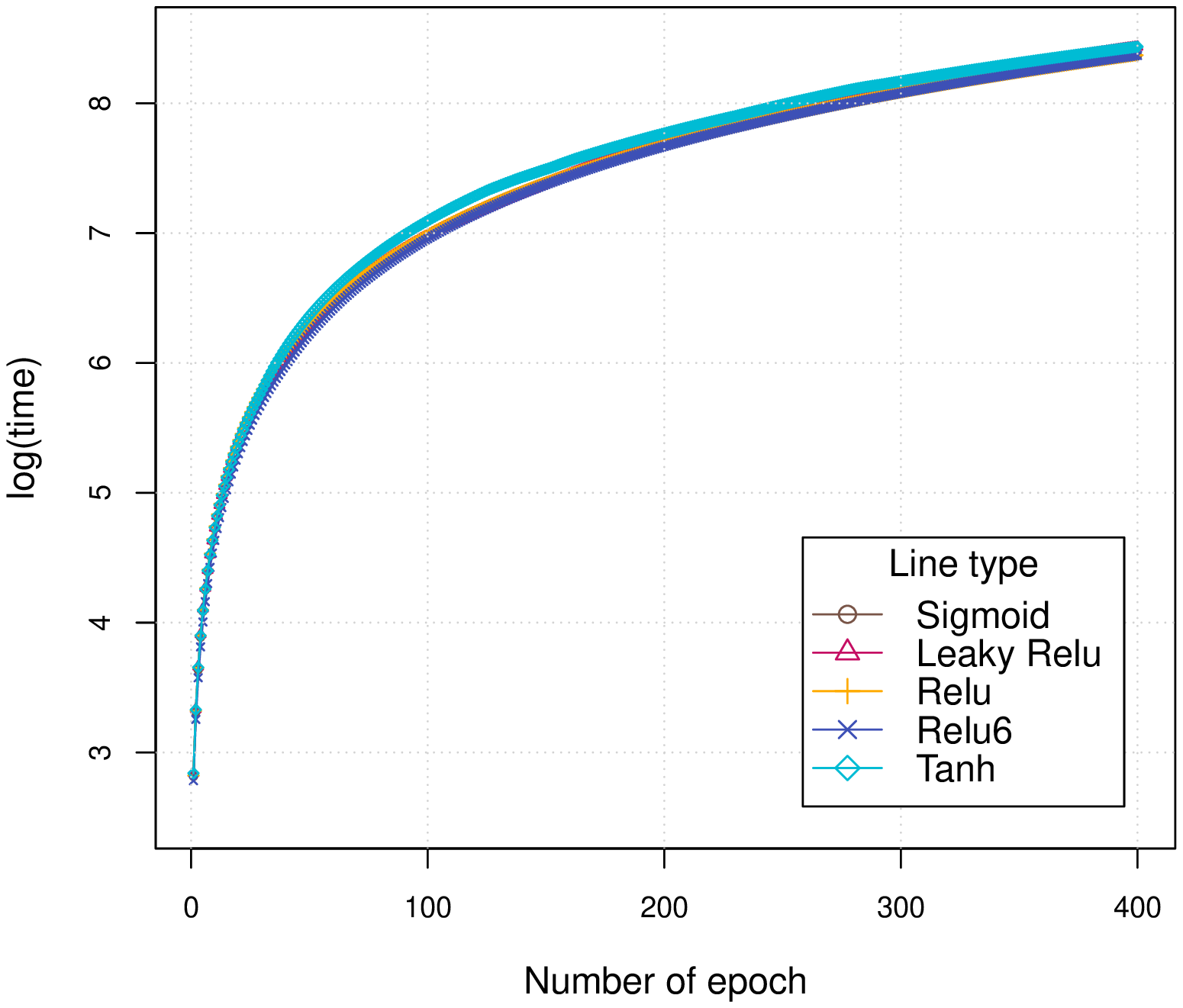}
\centering{(e)}
\centering
\end{minipage}
\begin{minipage}[t]{5.7cm}
\includegraphics[width=\textwidth]{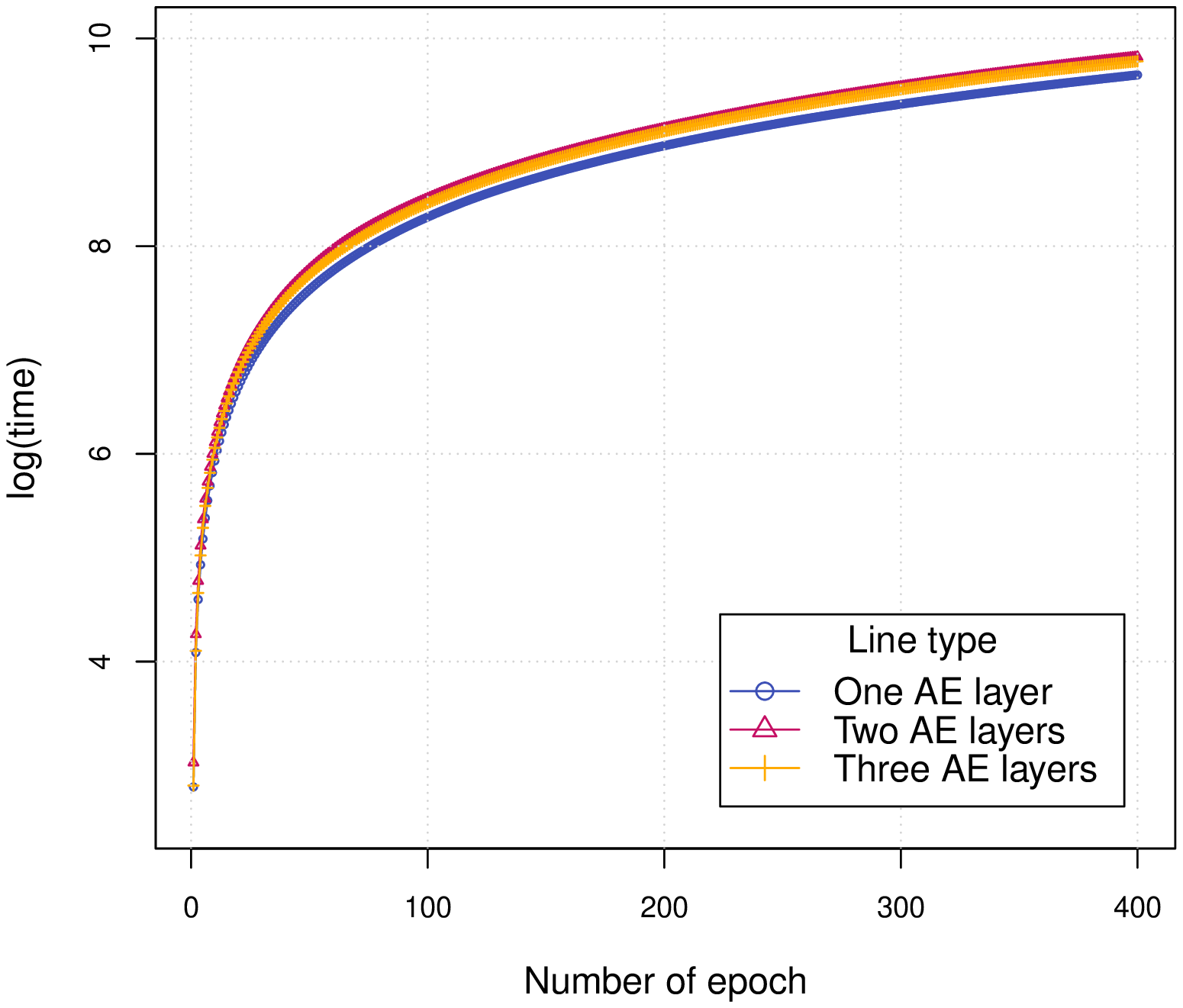}
\centering{(f)}
\centering
\end{minipage}
\begin{minipage}[t]{5.7cm}
\includegraphics[width=\textwidth]{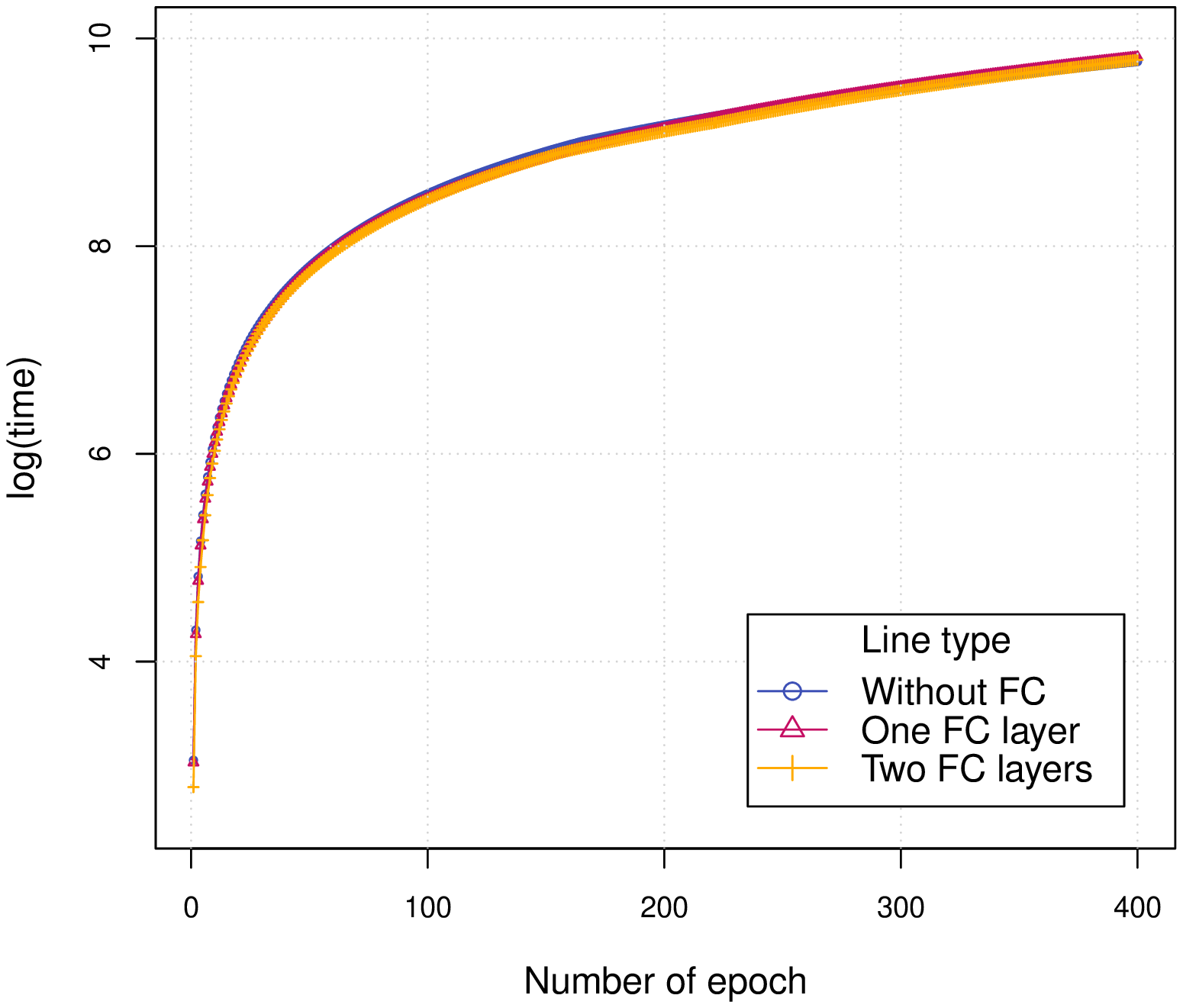}
\centering{(g)}
\centering
\end{minipage}
\begin{minipage}[t]{5.7cm}
\includegraphics[width=\textwidth]{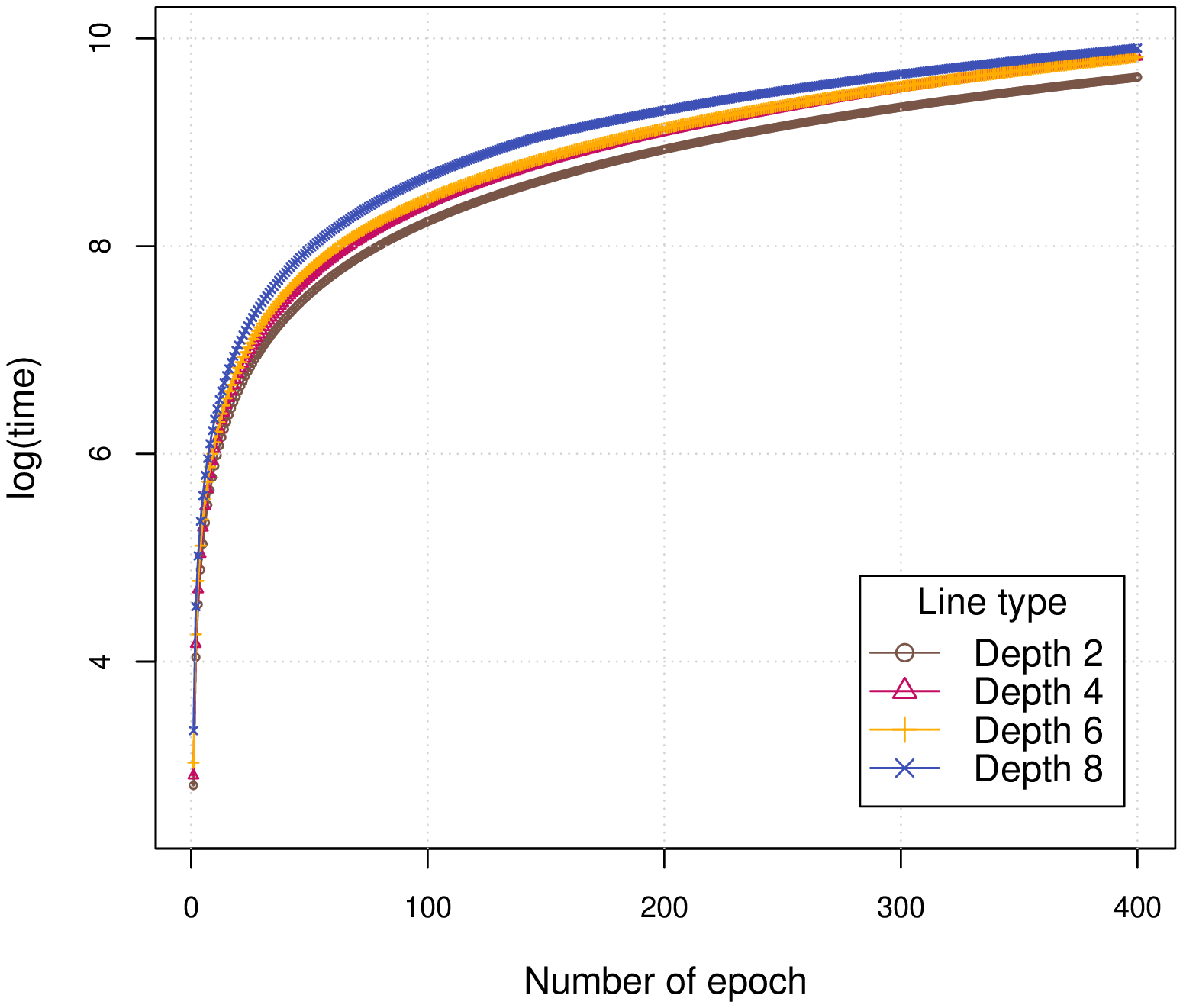}
\centering{(h)}
\centering
\end{minipage}
\begin{minipage}[t]{6.1cm}
\includegraphics[width=5.8cm]{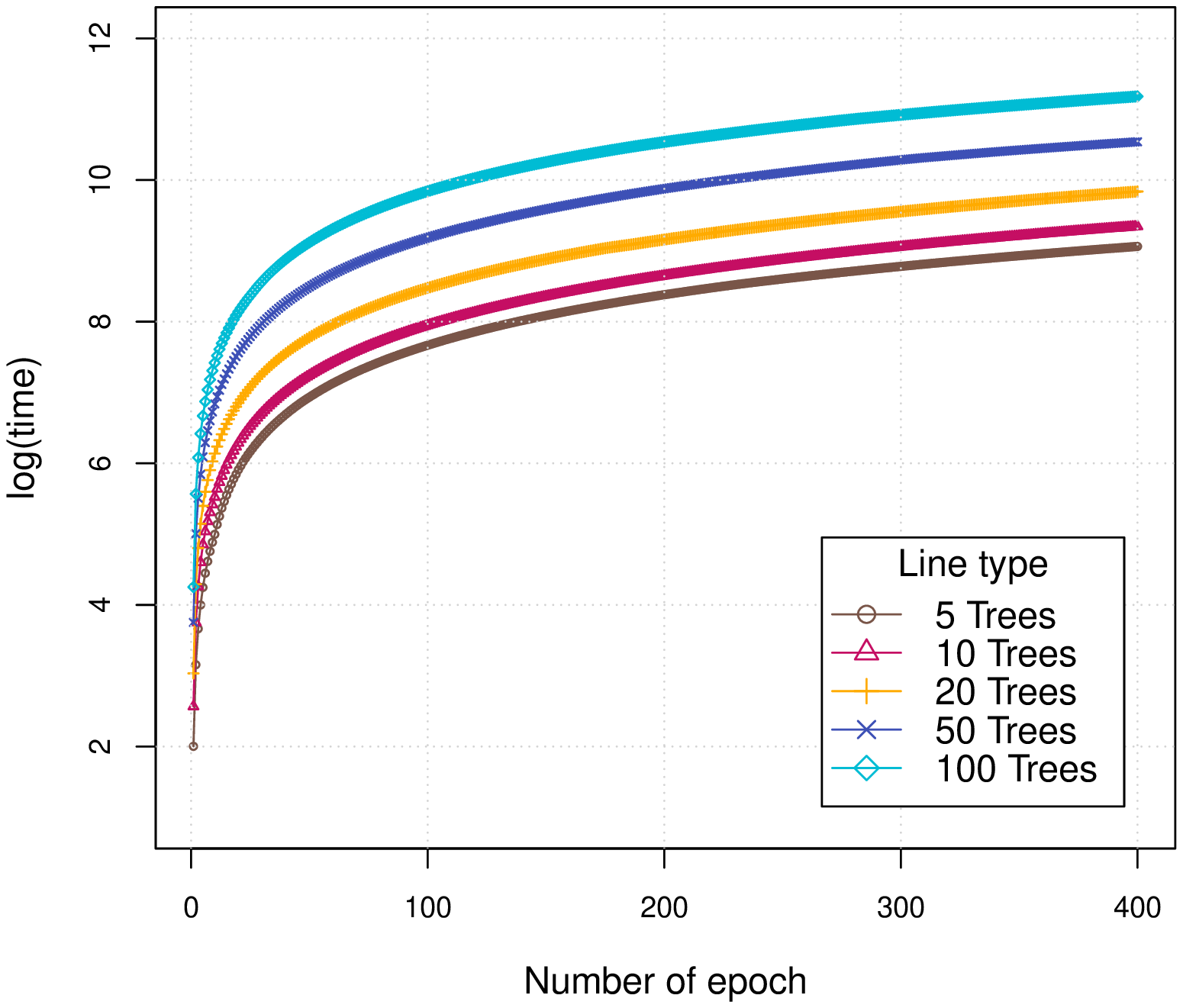}
\centering{(g)}
\centering
\end{minipage}
\begin{minipage}[t]{6.1cm}
\includegraphics[width=5.8cm]{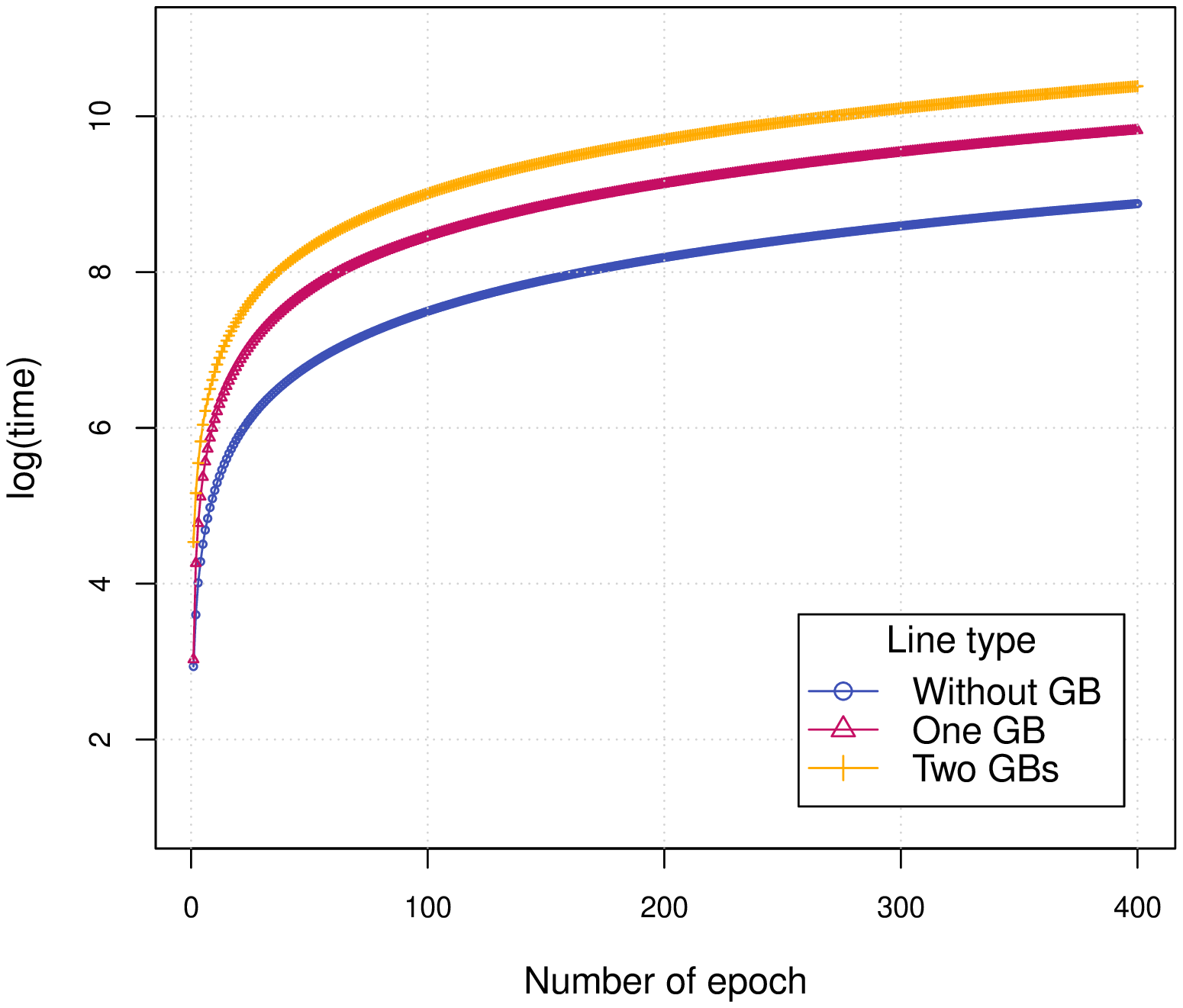}
\centering{(h)}
\centering
\end{minipage}
\caption{The accuracy for 400 epochs of the parameter: (a)the activation function type, (b)the number of autoencoder layers, (c)the number of fully connected layers, (d)the number of depth, (e)the number of trees, (f)the number of gradient boost module.}
\label{fig:accuracy}
\end{figure*}

To help fully understand and analyze the properties of our model, we further study the following key parameters with different settings:
\begin{itemize}
    \item \textbf{The activation function type}: the activation function type of autoencoder and fully connected layers, will be either Sigmoid function, Relu, leaky Relu, Relu6, or Tanh. 
    \item \textbf{The number of convolutional layers in convolutional autoencoders}: from one to three.  
    \item \textbf{The number of fully connected layers}: none, one layer, or two layers. 
    \item \textbf{The depth of the tree in each neural decision tree}: from 2 to 8.
    \item \textbf{Number of trees in a forest}: from 5 to 100.  
    \item \textbf{Number of the gradient boost modules}: none, one gradient boost module, or two gradient boost modules. 
\end{itemize}

We tune these parameters on the Fashion-MNIST dataset, and the default base learner is with two convolutional autoencoder layers in the initial and followed with one fully connected layers, then feed the outputs of a fully connected layer into a neural decision forest with 20 trees and depth as 3. The number of gradient boost module is one. And we controlled the parameters in the deep neural network's layers within 10 times of the variable size. 

We compared the parameters in both accuracy and time cost. The results are shown in Figure \ref{fig:accuracy}. From the accuracy lines, we could see that the
activation function has a limited impact on the results, and among the five activation functions, the accuracy of leaky relu activation function is lower than others. 
A model performs better with a two-layers convolutional autoencoder than a one-layer autoencoder in most time, and one layer convolutional autoencoder is easier to transfer the information of the inputs thus it performs the best in the first few iterations. Then we could notice a three-layer autoencoder has slowest converge rate since it needs more time to capture the complex relationship between inputs and the prediction.
And a model without fully connected layers will rapidly converge to an acceptable accuracy level. In the meanwhile, with the increase of the number of layers, it takes more epochs for a model to a convergence. 
We could also note that accuracy will quickly increase with a larger number of depth and trees, but it needs more iterations for learning with a small number of trees in the forest.
Both models with one gradient boost module and two gradient boost modules perform better than a model without gradient boost module, which indicates the efficiency of our models.

Figure~\ref{fig:accuracy} (e) to (h) show the time consumption under different parameter settings, where the axis y shows the $log$ of seconds. We observed the activation function type has only a slight effect on the time consumption. The time cost also increased slowly with the increase of the number of depth, the number of trees, and the number of gradient boost modules.

Considering the time cost and the prediction performance, we chose the sigmoid function as our activate function and two layers in the convolutional autoencoder as the default setting of our proposed model. That includes one fully connected layer of 50 trees in a neural decision forest with the depth of 5, and one gradient boost module. We defined the batch size as 200 and took min-max normalization method for each pixel. The best accuracy for Fashion-MNIST dataset is above 92\%.

\section{Conclusion}
Neural decision forest is a model which exploits the benefits of both random forest and deep neural networks. In this paper, we design a gradient boost module that is suitable for neural decision forest for better classification. We first introduce the framework of our models, including the convolutional autoencoder layers and fully connected layers as the prepossessing components, followed by the neural decision forest and the gradient boost module. We have applied our model with a series public datasets and the experimental results demonstrate the efficacy of our model over various learning tasks. 

\ifCLASSOPTIONcaptionsoff
  \newpage
\fi

\end{document}